%% file: main.tex
\documentclass{article}

\usepackage[preprint]{neurips_2026}
\usepackage[utf8]{inputenc} % allow utf-8 input
\usepackage[T1]{fontenc}    % use 8-bit T1 fonts
\usepackage{hyperref}       % hyperlinks
\usepackage{url}            % simple URL typesetting
\usepackage{booktabs}       % professional-quality tables
\usepackage{amsfonts}       % blackboard math symbols
\usepackage{amsmath}        % math
\usepackage{amssymb}        % math symbols
\usepackage{nicefrac}       % compact symbols for 1/2, etc.
\usepackage{microtype}      % microtypography
\usepackage{xcolor}         % colors
\usepackage{graphicx}       % figures

\graphicspath{{images/}}

\title{MaSC: A Masked Similarity Metric for Evaluating Concept-Driven Generation}

\title{MaSC: A Masked Similarity Metric for Evaluating Concept-Driven Generation}

\author{%
  Patryk Bartkowiak \\
  Adam Mickiewicz University \\
  \And
  Lennart Petersen \\
  Kiel University \\
  \And
  Bartosz Kotrys \\
  ArtCollect \\
  \And
  Dominik Michels \\
  KAUST \\
  \AND
  Soren Pirk \\
  Kiel University \\
  \And
  Wojtek Palubicki \\
  Adam Mickiewicz University \\
}

\begin{document}

\maketitle

\input{src/abstract.tex}

\input{src/introduction.tex}

\input{src/related.tex}

\input{src/method.tex}

\input{src/results.tex}

\input{src/discussion.tex}

\input{src/conclusion.tex}

\begin{ack}
Use unnumbered first level headings for the acknowledgments. All acknowledgments
go at the end of the paper before the list of references. Moreover, you are required to declare
funding (financial activities supporting the submitted work) and competing interests (related financial activities outside the submitted work).
More information about this disclosure can be found at: \url{https://neurips.cc/Conferences/2026/PaperInformation/FundingDisclosure}.

Do {\bf not} include this section in the anonymized submission, only in the final paper. You can use the \texttt{ack} environment provided in the style file to automatically hide this section in the anonymized submission.
\end{ack}

% \section*{References}

% References follow the acknowledgments in the camera-ready paper. Use unnumbered first-level heading for
% the references. Any choice of citation style is acceptable as long as you are
% consistent. It is permissible to reduce the font size to \verb+small+ (9 point)
% when listing the references.
% Note that the Reference section does not count towards the page limit.
% \medskip

% {
% \small

% [1] Alexander, J.A.\ \& Mozer, M.C.\ (1995) Template-based algorithms for
% connectionist rule extraction. In G.\ Tesauro, D.S.\ Touretzky and T.K.\ Leen
% (eds.), {\it Advances in Neural Information Processing Systems 7},
% pp.\ 609--616. Cambridge, MA: MIT Press.
% }

%%%%%%%%%%%%%%%%%%%%%%%%%%%%%%%%%%%%%%%%%%%%%%%%%%%%%%%%%%%%

%\bibliographystyle{plainnat}
\bibliographystyle{plain}
\bibliography{main}

%%%%%%%%%%%%%%%%%%%%%%%%%%%%%%%%%%%%%%%%%%%%%%%%%%%%%%%%%%%%

\input{src/appendix}
\clearpage

%%%%%%%%%%%%%%%%%%%%%%%%%%%%%%%%%%%%%%%%%%%%%%%%%%%%%%%%%%%%

\newpage
\input{checklist.tex}

\end{document}

%% file: src/abstract.tex
\begin{abstract}
Evaluating single-concept personalization in text-to-image diffusion has seen two categories of quantitative metrics: Concept Preservation (CP) measures identity fidelity to a reference while Prompt Following (PF) measures whether the generated scene matches the prompt. Personalization papers have commonly computed these signals using three separate backbones: CLIP-I and DINO for CP, CLIP-T for PF. In this paper, we show that existing metrics fall short of correlating with human perception because they attend to the image as a whole, instead of distinguishing the concept subject from the background. This distinction is important from a human perception point of view as the concept subject in the output image should be very similar to the input concept (CP), whereas the output background should adhere closely to the text prompt (PF). To improve personalization evaluation in this way, we introduce \emph{MaSC}, a unified metric that attends to CP and PF by differentiating concept subject regions. Specifically, given an externally provided foreground concept mask, MaSC computes the CP and PF scores from a single forward pass of a frozen SigLIP2 encoder per image. 
%The CP score is a masked max-cosine over patch tokens, restricted to the foreground concept region of the reference. The PF score reuses the same patch tokens, passing them through SigLIP2's trained attention pooler with foreground patches masked from the attention computation, then taking cosine against a prompt embedding where the subject has been stripped off. 
On DreamBench++ human ratings, MaSC reaches Krippendorff $\alpha = +0.471$ on CP --- beating every non-LLM baseline tested (DINOv3, DreamSim, AM-RADIO, DIFT-SDXL, DINO-I, CLIP-I) and GPT-4V, while sitting within $\Delta\alpha = 0.028$ of GPT-4o. To distinguish human perception bias we also evaluate our method, as well as common SOTA baselines, on ORIDa, a real-photo benchmark of identity preservation across physical environments. In this experiment MaSC reaches AUC $= 0.992$ almost perfectly identifying concept subjects. The PF score, obtained by MaSC without a second encoder forward pass, beats the CLIP-T baseline shipped with DreamBench++. In summary, our comprehensive evaluations demonstrate that MaSC establishes a new state-of-the-art for non-LLM concept preservation, while providing an efficient, unified standard for personalization evaluation. We release MaSC as a pip-installable Python package, alongside independent reproductions of every comparator's published numbers. 
\end{abstract}

%% file: src/introduction.tex
\section{Introduction}
\label{sec:introduction}

The ability to personalize text-to-image (T2I) generation to user-supplied subjects has fundamentally transformed generative AI. Methods such as DreamBooth~\cite{ruiz2022dreambooth}, Textual Inversion~\cite{gal2022imageworthwordpersonalizing}, and IP-Adapter~\cite{ye2023ip-adapter} have made subject-driven generation a core research pillar and a standard T2I capability. However, reliably evaluating the success of these personalizations remains a difficult task. A complete evaluation requires capturing two distinct signals: \emph{Concept Preservation} (CP), measuring the identity fidelity of the rendered subject to its reference, and \emph{Prompt Following} (PF), measuring how well the generated scene satisfies the prompt. In the past, the field computed these signals using three independent backbones inherited from DreamBooth: CLIP-I~\cite{radford2021learningtransferablevisualmodels} and DINO~\cite{9709990} for CP, and CLIP-T for PF. Recently, DreamBench++~\cite{peng2024dreambench} exposed the severe limitations of this standard trio, revealing that its correlation with human judgments plateaus at a Krippendorff~\cite{krippendorff} $\alpha \approx 0.3$ on both signals—roughly half the inter-annotator ceiling ($\alpha = 0.658$ for CP, $0.563$ for PF). To improve alignment with human judgments, the community has generally taken two approaches: updating the representation encoders, or using Large-Language Models (LLMs) like GPT-4V~\cite{openai2024gpt4technicalreport} and GPT-4o~\cite{openai2024gpt4ocard} as evaluators. The first approach offers only minor improvements, while the second performs well but relies on API-bound, non-differentiable scoring that lacks per-region attribution.

The plateau is not a limitation of the encoders. Global pooling introduces a distinct failure for each evaluation signal: (i) Global cosine for CP averages over the whole image, including background variation that humans correctly ignore when judging identity; (ii) Global text-image cosine for PF combines two distinct quantities: scene adherence to the prompt and the presence of the subject class (a static factor, since the subject is fixed across all test prompts for a given concept). Both failures are consequences of the aggregator and the input space, not the backbone --- any encoder under mean or [CLS] pooling inherits them.

These limitations are addressable at inference time, without retraining, via spatial decomposition. We introduce \emph{MaSC}, which uses three lightweight interventions: (i) for each foreground reference patch, taking the maximum cosine against any output patch and averaging under the foreground mask; (ii) applying SigLIP2's trained attention pooler over the patch tokens with foreground patches masked, yielding a background-only embedding via the pooler's native masking support; and (iii) excluding the canonical subject name from the prompt, eliminating the static baseline on the text side. A single SigLIP2 SO400M-NaFlex~\cite{tschannen2025siglip} forward pass per image then produces both scores: CP from the patch grid via masked max-cosine, and PF from the masked attention pool against the stripped-prompt text embedding.

MaSC's spatial decomposition closes most of the performance gap with LLM judges in personalization evaluation. On DreamBench++, MaSC is the only non-LLM metric to outperform GPT-4V on CP. On ORIDa~\cite{kim2025orida}, a real-photo identity-discrimination benchmark across 50 subjects in 10 environments, MaSC is the first non-LLM metric to surpass GPT-4o. Furthermore, the PF score outperforms the CLIP-T baseline shipped with DreamBench++.

\paragraph{Contributions.} We propose MaSC, a unified personalization metric that produces CP and PF scores from a single SigLIP2 SO400M-NaFlex forward pass per image, given an externally supplied foreground concept mask. We make five contributions: (1) On DreamBench++ human ratings, MaSC achieves Krippendorff $\alpha = +0.471$ on CP --- outperforming every non-LLM baseline tested, surpassing GPT-4V, and reaching $72\%$ of the human inter-rater ceiling; (2) On ORIDa, MaSC is the first non-LLM CP metric to outscore GPT-4o, reaching AUC $= 0.992$; (3) Same-backbone ablations isolate the contribution to our spatial decomposition strategy rather than the encoder choice, with this approach accounting for $\Delta\alpha = +0.102$ on DreamBench++ and $\Delta\,\mathrm{AUC} = +0.038$ on ORIDa; (4) At a matched parameter budget, MaSC outperforms state-of-the-art distilled vision transformers (e.g., AM-RADIO~\cite{Ranzinger_2024_CVPR}), demonstrating that late-stage explicit masking combined with patch features consistently outperforms internalized mask supervision for personalization evaluation; (5) The PF score, obtained from the same forward pass without an additional vision-encoder call, beats the CLIP-T baseline shipped with DreamBench++.

%% file: src/related.tex
\section{Related Work}
\label{sec:related}

Subject-driven text-to-image personalization spans optimization-based methods that fine-tune or augment the generator with a reference subject (DreamBooth, Textual Inversion) and adapter-based methods that inject reference features at inference (IP-Adapter, BLIP-Diffusion~\cite{li2023blip}, Emu2~\cite{Emu2}). While these generative techniques form the core of the personalization literature, our work focuses strictly on the evaluation metrics used to benchmark them.
DreamBooth introduced the evaluation protocol of using CLIP-I and DINO-I for image-image fidelity alongside CLIP-T for image-text adherence. This protocol has since been universally adopted. However, DreamBench++~\cite{peng2024dreambench} recently benchmarked these metrics against human judgments across an evaluation grid of 7 personalization methods, 150 subjects, and 9 prompts. They demonstrated that with CLIP-I at Krippendorff $\alpha = +0.135$, DINO-I and CLIP-T plateau near $\alpha \approx 0.3$, roughly half the human inter-annotator ceiling ($\alpha = 0.658$ for CP, $0.563$ for PF). Our work directly addresses this critical evaluation gap. To complement DreamBench++, we also evaluate on ORIDa, a real-photo benchmark of identity preservation across natural environments. While ORIDa lacks human ratings, its inclusion of ground-truth per-object segmentation masks makes it an ideal real-world out-of-distribution test for CP.
Several non-LLM alternatives have been proposed for image-image fidelity beyond CLIP-I and DINO-I. DreamSim~\cite{fu2023dreamsim} integrates CLIP, DINO, and OpenCLIP backbones and fine-tunes on the NIGHTS triplet dataset for perceptual similarity. DINOv3~\cite{simeoni2025dinov3} is a modern self-supervised vision transformer that updates the original DINO-I baseline. AM-RADIO C-RADIOv4-SO400M is a ViT trained via distillation from SigLIP2-g, DINOv3-7B, and SAM3~\cite{carion2025sam3segmentconcepts} simultaneously, internalizing within the encoder weights what MaSC supplies externally as a mask. DIFT~\cite{tang2023emergent} extracts dense correspondence features from frozen Stable Diffusion~\cite{rombach2022highresolutionimagesynthesislatent} U-Nets. While these methods seek to improve representation through training or architecture, MaSC explores a complementary direction: decomposing the image into concept-specific regions at inference time to isolate identity-relevant features.
To address the performance limit of non-LLM metrics, multimodal LLMs have been increasingly employed as judges. DreamBench++ evaluates GPT-4V~\cite{openai2024gpt4technicalreport} and GPT-4o~\cite{openai2024gpt4ocard}  on its human-rated subset; GPT-4o achieves CP $\alpha = +0.499$, approaching the human inter-annotator ceiling. Despite their high correlation with humans, these judges are API-bound, non-differentiable, and lack per-region attribution: their scores are scalar outputs that provide no spatial grounding for the judgment. We benchmark against GPT-4V and GPT-4o as the canonical LLM-judge references in the personalization literature.
On the PF side, dedicated reward models provide an alternative to LLM judges. ImageReward~\cite{xu2023imagereward} utilizes a BLIP~\cite{li2022blipbootstrappinglanguageimagepretraining} backbone fine-tuned on pairwise human preferences; HPSv3~\cite{ma2025hpsv3widespectrumhumanpreference} extends this paradigm with a Qwen2-VL~\cite{wang2024qwen2vlenhancingvisionlanguagemodels} architecture. VQAScore~\cite{lin2024evaluating} queries a CLIP-FlanT5 model with `Does this figure show {prompt}?' and uses the probability of a 'Yes' answer as the fidelity score. These are dedicated PF reward models trained on human-preference data. VQAScore ($\sim$3B) and HPSv3 ($\sim$8B) wrap larger backbones, while ImageReward ($\sim$360M) is more compact. Two of the three outperform MaSC on DreamBench++ PF (VQAScore $\alpha = +0.504$, ImageReward $\alpha = +0.441$), while HPSv3 ($\alpha = +0.299$) underperforms MaSC ($\alpha = +0.354$) despite its larger parameter budget. We position the MaSC PF score as a secondary signal obtained from the primary CP forward pass. It outperforms the standard CLIP-T baseline without requiring additional vision-encoder inference or dedicated reward-model training.

%% file: src/method.tex
\section{Method}
\label{sec:method}

\begin{figure*}[t]
    \centering
    \includegraphics[width=\textwidth]{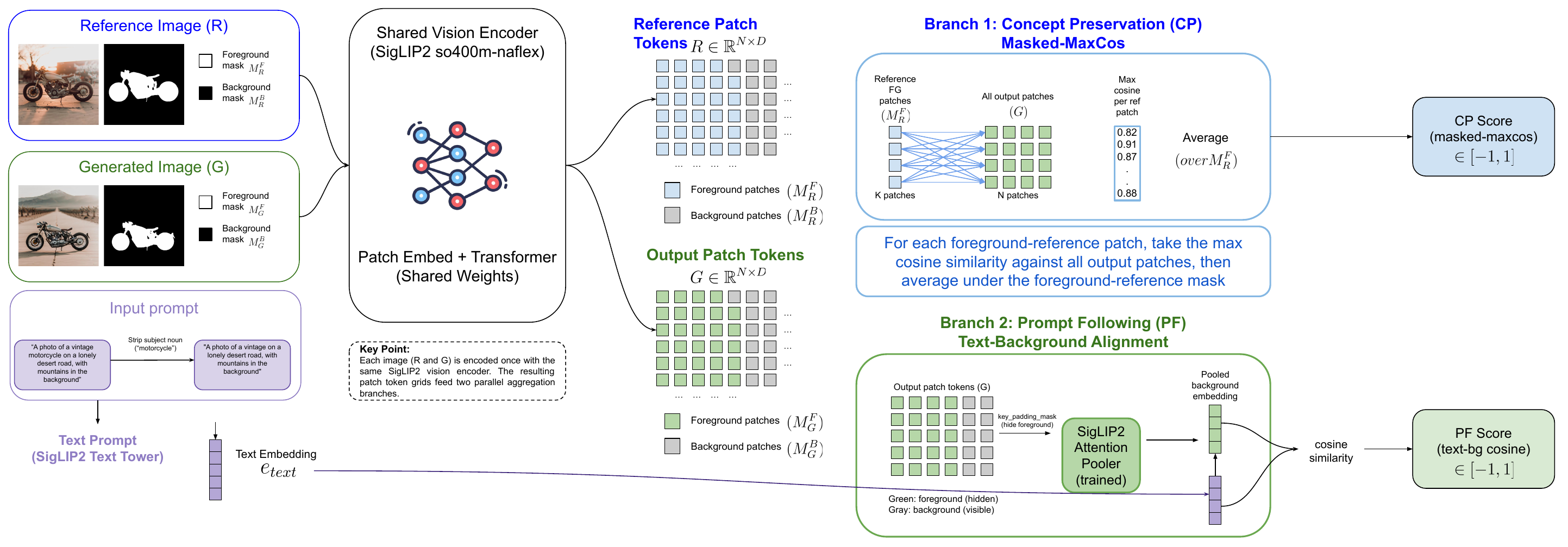}
    \caption{\textbf{MaSC pipeline overview.} Reference and generated images are encoded once each by a frozen SigLIP2 SO400M-NaFlex vision tower, producing patch-token grids $R, G \in \mathbb{R}^{N \times D}$. MaSC consumes provided foreground concept masks, denoted $M_R^F, M_G^F \in \{0, 1\}^{H \times W}$. Two branches share the single forward pass. \textit{Branch 1 --- Concept Preservation (masked-maxcos):} for each patch covered by the reference foreground mask ($M_R^F$), take its maximum cosine similarity against all output patches in $G$, then average over the foreground-reference patches. \textit{Branch 2 --- Prompt Following (Text-Background Alignment):} the trained SigLIP2 attention pooler is run with foreground patches of $G$ masked from the attention computation, producing a background-only embedding. This is compared via cosine similarity to the text embedding $e_{text}$ of the prompt with the subject noun stripped.}
    \label{fig:pipeline}
    \vspace{-4mm}
\end{figure*}

We now describe MaSC, a unified evaluation metric that uses explicit concept masks to decompose each generated image into subject and background regions, enabling concept preservation and prompt following to be scored from shared frozen SigLIP2 features (Figure~\ref{fig:pipeline}).

\subsection{Setup and Notation}
\label{sec:method:setup}

We are given a reference image $\mathcal{I}_R$, an output image $\mathcal{I}_G$ generated by a personalization method, the natural-language prompt $p$ used to condition the generation, and the subject's canonical name $w$ (e.g.\ ``kitten'', ``piggy bank''). MaSC utilizes explicitly provided concept masks $M_R, M_G \in \{0, 1\}^{H \times W}$ for the two images.

A frozen SigLIP2 SO400M-NaFlex vision tower encodes each image once. The vision tower outputs $N$ patch tokens in $\mathbb{R}^D$ at the NaFlex max-patch budget ($N = 1024$ for our square inputs); we write these as matrices $R, G \in \mathbb{R}^{N \times D}$ and let $\tilde{r}_i, \tilde{g}_j$ denote the $\ell_2$-normalized tokens. The text tower $T_\theta$ produces a single embedding in the joint image-text space. The trained image embedding head $\mathrm{Pool}_\theta(\cdot)$ is a single learned-query attention pool over the patch tokens; critically, the pooler accepts a binary suppression mask that excludes arbitrary positions from the attention computation, without changing the trained parameters.

Each $H \times W$ image-resolution mask is downsampled to the patch grid via bilinear interpolation and thresholded at $0.5$ to obtain patch-level binary masks $m_R, m_G \in \{0, 1\}^N$. The two scoring branches below share the vision-tower output of each image.

\paragraph{Mask source and filtering.} The metric itself is mask-source-agnostic. For our DreamBench++ experiments, we extract $M_R$ and $M_G$ with SAM3 prompted by the canonical concept name $w$, applied identically to reference and output images. For ORIDa, we use the per-object segmentation masks provided with the dataset. We discard pairs whose foreground masks are too small to carry reliable signal --- on DreamBench++, where masks are extracted by SAM3, a near-empty foreground mask typically reflects either a segmentation failure on the reference or a generated output that did not render the subject. This filtering removes degenerate cases where the subject is effectively absent or localization fails, and is applied uniformly across all evaluated methods, independent of the scoring metric.  We use a $5\%$ of the image area threshold on DreamBench++: when an output mask falls below $5\%$, that pair is dropped; when a reference mask falls below $5\%$, all $9$ prompt variants for that subject are dropped, since the reference is reused across them and the failure is intrinsic to the reference rather than to a single output. On ORIDa, the threshold is $0.5\%$, lower because the provided masks are reliable and the dataset's real-world objects span a wider scale range than DreamBench++'s synthetic subjects. The PF leaderboard additionally excludes the $20$ DreamBench++ ``style'' subjects, whose prompts describe a visual style rather than a scene; subject-stripping on these prompts produces grammatically malformed fragments. Any segmentation method producing $\{0, 1\}^{H \times W}$ outputs can be substituted for SAM3 without retraining.

\subsection{Concept Preservation: \texttt{masked\_maxcos}}
\label{sec:method:cp}

Let $\mathcal{F}_R = \{i : m_R^{(i)} = 1\}$ index the foreground-reference patches. For each $i \in \mathcal{F}_R$, we take the maximum cosine similarity against any output patch:
\[
    s_i = \max_{j \in \{1, \ldots, N\}} \langle \tilde{r}_i, \tilde{g}_j \rangle.
\]
The CP score is the mean of those per-patch best matches under the foreground-reference mask:
\begin{equation}
\label{eq:cp}
    \mathrm{CP}(R, G, m_R) = \frac{1}{|\mathcal{F}_R|} \sum_{i \in \mathcal{F}_R} s_i.
\end{equation}

Three design choices are encoded in Equation~\eqref{eq:cp}. First, the mask is applied only on the reference side; the output is searched in full, ensuring the metric does not penalize spatial relocation of the concept. Second, the max-cosine inherently measures whether each local patch of the reference concept is present anywhere in the output. This approach degrades gracefully under partial identity preservation, unlike mutual-nearest-neighbor matching, whose strict one-to-one constraint severely over-penalizes the partial-preservation regime typical of personalization outputs. Third, averaging over the foreground mask aggregates the patch-level matches without overweighting smaller concept regions. As demonstrated in Section~\ref{sec:results:aggregator}, substituting Equation~\eqref{eq:cp} with mutual-nearest-neighbor matching (using match-count normalization on identical features) drops the Krippendorff $\alpha$ by $0.462$ --- an aggregation failure that outweighs the contribution of the encoder backbone itself.

\subsection{Prompt Following: Background-pool and subject stripping}
\label{sec:method:pf}

The PF score is derived via two targeted interventions, one per modality.

\paragraph{Image side: background-only attention pooling.} We apply the trained attention pooler to the generated image $G$ while masking its foreground patches from the attention computation, restricting the pool to background tokens only:
\[
    e_G^{\mathrm{bg}} = \mathrm{Pool}_\theta(G;\, m_G).
\]
Here, $m_G$ serves as a binary suppression mask --- positions where $m_G^{(i)} = 1$ are explicitly excluded from the pool. The pooler operates with frozen parameters, only the visibility of the input tokens is altered. Consequently, the output remains in the joint image-text space, enabling direct cosine comparison with text embeddings.

\paragraph{Text side: subject-name stripping.} Given the canonical subject name $w$, we strip $w$ (and any leading articles) from the prompt $p$ using case-insensitive regular expression matching. This string-matching robustly handles multi-word and hyphenated names. Excess whitespace is subsequently collapsed, yielding the stripped prompt $p' = \mathrm{strip}(p, w)$.

\paragraph{Score.}
\begin{equation}
\label{eq:pf}
    \mathrm{PF}(G, m_G, p, w) = \langle \tilde{e}_G^{\mathrm{bg}},\, \tilde{T}_\theta(p') \rangle.
\end{equation}

The intuition: In personalization evaluation, the subject is fixed across all prompts for a given concept. Consequently, whole-image text-image cosine similarity fails to separate the desired signal (scene-prompt alignment) from a persistent baseline. Hiding the foreground from the image pooler suppresses this baseline on the visual side, while stripping the subject name removes the corresponding expectation from the text side. As demonstrated in Section~\ref{sec:results:pf}, a foreground-pooled control yields a negative Krippendorff $\alpha$, confirming that information strictly contained within the foreground actively degrades prompt-following evaluation.

\subsection{Compute Profile}
\label{sec:method:compute}

To evaluate a given $(\mathcal{I}_R, \mathcal{I}_G, p, w)$ tuple, MaSC requires: two vision-tower forward passes (one per image, utilizing the $\sim$428M parameter SigLIP2 SO400M-NaFlex encoder), a single execution of the trained attention pooler on the cached output patches with foreground positions suppressed via $m_G$, and one text-tower forward pass on the stripped prompt. Because the CP branch directly reuses the cached vision features, computing both the CP and PF metrics jointly incurs no additional vision-encoder cost beyond the initial feature extraction.

%% file: src/results.tex
\section{Results}
\label{sec:results}

We evaluate MaSC on human-rated personalization benchmarks and real-photo identity-discrimination tasks, comparing it against standard CLIP/DINO metrics, modern non-LLM similarity models, reward-based prompt-following scores, and multimodal LLM judges.

\subsection{CP on DreamBench++}
\label{sec:results:cp:dbplus}

Table~\ref{tab:cp_dbplus} reports CP $\alpha$ and $\rho$ for MaSC and ten comparators on the $7{,}135$-key DreamBench++ subset --- the strictly matched intersection of all standard baselines, both LLM judges, and the new comparator runs.
MaSC is the only non-LLM metric to outperform GPT-4V on the benchmark. It trails GPT-4o (the only superior entry) by just $\Delta\alpha = 0.028$, reaching $72\%$ of the human inter-annotator ceiling.

\begin{table}[t]
    \centering
    \caption{Concept Preservation on DreamBench++. Krippendorff $\alpha$ (Kd$_o$) and Spearman $\rho$ against pooled human ratings on the $7{,}135$-key apples-to-apples subset. \textbf{Bold}: ours --- the only non-LLM, non-human metric to beat GPT-4V; sits $\Delta\alpha = -0.028$ behind GPT-4o and reaches $72\%$ of the honest human inter-rater ceiling.}
    \label{tab:cp_dbplus}
    \scalebox{0.8}{
    \begin{tabular}{rlcrr}
        \toprule
        \# & Metric & Kind & $\alpha$ & $\rho$ \\
        \midrule
        --- & \emph{Human inter-rater} & \emph{ceiling} & \emph{+0.658} & \emph{+0.653} \\
        1 & GPT-4o & LLM judge & +0.499 & +0.641 \\
        2 & \textbf{MaSC} & \textbf{non-LLM} & \textbf{+0.471} & \textbf{+0.611} \\
        3 & GPT-4V & LLM judge & +0.432 & +0.521 \\
        4 & DreamSim & non-LLM & +0.421 & +0.599 \\
        5 & SigLIP2 SO400M-NaFlex global pool & non-LLM & +0.369 & +0.557 \\
        6 & DINOv3 & non-LLM & +0.345 & +0.479 \\
        7 & DIFT-SDXL & non-LLM & +0.324 & +0.502 \\
        8 & DINO-I & non-LLM & +0.311 & +0.514 \\
        9 & AM-RADIO C-RADIOv4-SO400M & non-LLM & +0.226 & +0.541 \\
        10 & CLIP-I & non-LLM & +0.135 & +0.533 \\
        \bottomrule
    \end{tabular}
    }
\end{table}

\paragraph{Same-backbone ablation.} The same-backbone ablation establishes the contribution of the masked-maxcos aggregator as $\Delta\alpha = 0.102$: applying a global pool to the exact same SigLIP2 SO400M-NaFlex checkpoint yields $\alpha = +0.369$, confirming that the performance lift is driven purely by the aggregation strategy. Furthermore, while modern DINOv3 improves upon the standard DINO-I baseline ($\Delta\alpha = +0.034$), and DIFT-SDXL under its canonical recipe (timestep $261$, \texttt{up\_ft\_index=1}, null prompt) achieves $\alpha = +0.324$, both still trail MaSC by a margin exceeding $0.125\,\alpha$.

\paragraph{Architectural ablation at matched parameter budget.} We also compare against the two strongest architectural competitors in the modern non-LLM lineup: DreamSim (a NIGHTS-finetuned perceptual ensemble) and AM-RADIO C-RADIOv4-SO400M (which distills SigLIP2-g, DINOv3-7B, and SAM3 into a 431M ViT, matching our parameter budget and patch size). Both models underperform MaSC by $\Delta\alpha = 0.050$ and $\Delta\alpha = 0.245$, respectively.

\subsection{CP on ORIDa}
\label{sec:results:cp:orida}

\begin{figure}[t]
    \centering
    \includegraphics[width=\linewidth]{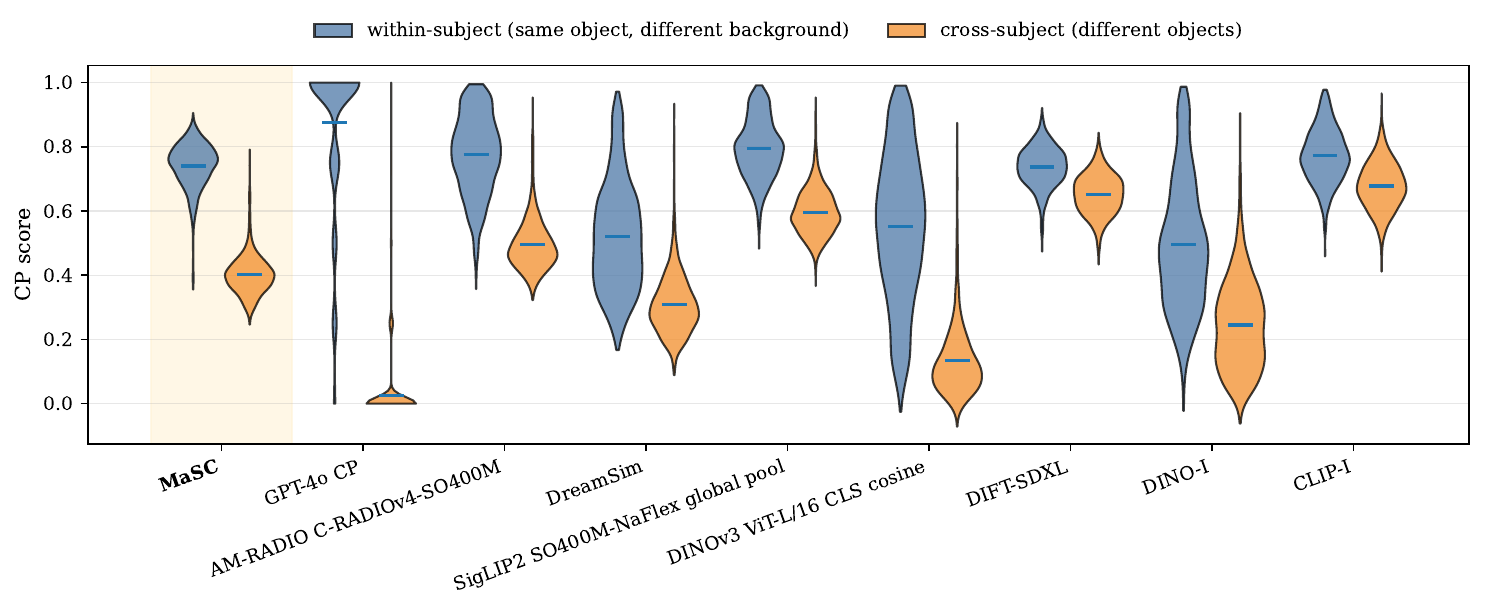}
    \caption{\textbf{ORIDa CP score distributions per metric:} within-subject pairs vs.\ cross-subject pairs; short horizontal lines mark the mean. Our proposed metric (MaSC) is highlighted with a shaded background. This figure visualizes the tail-overlap behavior that drives the AUC rankings in Table~\ref{tab:cp_orida}. Continuous-cosine metrics separate the two distributions cleanly, whereas the integer-rubric of GPT-4o yields the largest mean separation but heavily quantizes scores at rubric ties.}
    \label{fig:orida_violins}
\end{figure}

Table~\ref{tab:cp_orida} reports CP on ORIDa (train subset), a real-photo identity-discrimination benchmark containing ground-truth per-object segmentation masks. We sample $50$ subjects across $10$ backgrounds using $1$ photo per combination (specifically, the alphabetically first camera angle and first factual placement). From this, we form $50 \times \binom{10}{2} = 2{,}250$ within-subject cross-environment pairs and $2{,}250$ random cross-subject pairs. Because ORIDa lacks human ratings, we evaluate performance using the Area Under the Curve (AUC), representing the probability that a within-subject pair outscores a cross-subject pair: $\mathrm{AUC}(\text{within-pair} > \text{cross-pair})$. Figure~\ref{fig:orida_violins} visualizes the corresponding score distributions.

\begin{table}[t]
    \centering
    \caption{Concept Preservation on ORIDa. Real-photo identity discrimination across 50 subjects $\times$ 10 backgrounds: $2{,}250$ within-subject cross-environment pairs and $2{,}250$ random cross-subject pairs. $\mathrm{AUC}(\text{within} > \text{cross})$ is the probability a random within-subject pair scores higher than a random cross-subject pair. \textbf{Bold}: ours --- the first non-LLM CP metric to beat GPT-4o on a CP benchmark.}
    \label{tab:cp_orida}
    \scalebox{0.8}{
    \begin{tabular}{lrrrr}
        \toprule
        Metric & mean within & mean cross & $\Delta_\mathrm{norm}$ & AUC \\
        \midrule
        \textbf{MaSC} & $0.740 \pm 0.071$ & $0.402 \pm 0.064$ & $+0.513$ & \textbf{0.992} \\
        GPT-4o$^{*}$ & $0.875 \pm 0.242$ & $0.025 \pm 0.094$ & $+0.850$ & 0.986 \\
        AM-RADIO C-RADIOv4-SO400M & $0.775 \pm 0.124$ & $0.496 \pm 0.081$ & $+0.416$ & 0.961 \\
        SigLIP2 SO400M-NaFlex global pool & $0.795 \pm 0.092$ & $0.597 \pm 0.071$ & $+0.318$ & 0.954 \\
        DINOv3 & $0.553 \pm 0.230$ & $0.134 \pm 0.113$ & $+0.394$ & 0.942 \\
        DreamSim & $0.521 \pm 0.169$ & $0.308 \pm 0.093$ & $+0.241$ & 0.873 \\
        DINO-I & $0.495 \pm 0.201$ & $0.245 \pm 0.141$ & $+0.238$ & 0.848 \\
        DIFT-SDXL & $0.737 \pm 0.063$ & $0.652 \pm 0.062$ & $+0.174$ & 0.834 \\
        CLIP-I & $0.773 \pm 0.085$ & $0.678 \pm 0.074$ & $+0.167$ & 0.800 \\
        \bottomrule
    \end{tabular}
    }
    \par\smallskip
    {\footnotesize $^{*}$GPT-4o ratings rescaled from the native 0--4 rubric to $[0, 1]$ (raw $/\,4$) for cross-metric comparability. AUC is invariant under monotonic transformations.}
\end{table}

MaSC is the first non-LLM metric to outperform GPT-4o on a CP benchmark ($\Delta\,\mathrm{AUC} = 0.006$); this ranking inverts compared to DreamBench++ (where GPT-4o leads by $\Delta\alpha = 0.028$) as test conditions become more naturalistic. The underlying mechanism is visualized in Figure~\ref{fig:orida_violins}: continuous-cosine metrics cleanly separate the tails of the within- versus cross-subject distributions. In contrast, GPT-4o's discrete 0--4 rubric quantizes scores, inducing heavy ties. Consequently, while its mean separation is the largest evaluated, its AUC ceiling is artificially capped because numerous cross-subject pairs score at or above the minimum within-subject score. The same-backbone ablation extends to real-world photos: the global pool yields an $\mathrm{AUC}$ of $0.954$, whereas masked-maxcos achieves $0.992$. This $\Delta\,\mathrm{AUC} = 0.038$ improvement is smaller in absolute magnitude than the $\Delta\alpha = 0.102$ observed on DreamBench++, but consistent in direction. Notably, AM-RADIO outperforms DreamSim on ORIDa (reversing their DreamBench++ ranking), which aligns with AM-RADIO's distillation training successfully capturing real-world object semantics; conversely, DreamSim's NIGHTS-triplet fine-tuning transfers poorly to natural environment variations. DIFT-SDXL under its canonical recipe struggles again, yielding an $\mathrm{AUC}$ of $0.834$ (ranking penultimate). Ultimately, MaSC is the only metric to achieve top performance across both evaluation regimes.

\subsection{PF on DreamBench++}
\label{sec:results:pf}

Table~\ref{tab:pf_dbplus} reports PF $\alpha$ and $\rho$ on the $6{,}607$-key style-excluded DB++ subset. Style subjects are excluded because their prompts describe a visual style rather than a scene; consequently, subject-stripping these prompts produces grammatically malformed fragments.

\begin{table}[t]
    \centering
    \caption{Prompt Following on DreamBench++. Krippendorff $\alpha$ and Spearman $\rho$ against pooled human PF ratings on the $6{,}607$-key style-excluded subset. \textbf{Bold}: ours --- a free byproduct of the CP forward pass that beats CLIP-T on the same backbone by $\Delta\alpha = +0.027$. We do not claim PF SOTA; VQAScore and ImageReward lead the non-LLM lineup.}
    \label{tab:pf_dbplus}
    \scalebox{0.8}{
    \begin{tabular}{rlcrr}
        \toprule
        \# & Scorer & Kind & $\alpha$ & $\rho$ \\
        \midrule
        --- & \emph{Human inter-rater} & \emph{ceiling} & \emph{+0.563} & \emph{+0.571} \\
        1 & GPT-4o & LLM judge & +0.547 & +0.656 \\
        2 & VQAScore & non-LLM & +0.504 & +0.640 \\
        3 & GPT-4V & LLM judge & +0.485 & +0.661 \\
        4 & ImageReward & non-LLM & +0.441 & +0.611 \\
        5 & \textbf{MaSC} & \textbf{non-LLM} & \textbf{+0.354} & +0.538 \\
        6 & CLIP-T & non-LLM & +0.327 & +0.477 \\
        7 & SigLIP2 SO400M-NaFlex global pool & non-LLM & +0.326 & +0.466 \\
        8 & HPSv3 & non-LLM & +0.299 & +0.434 \\
        \bottomrule
    \end{tabular}
    }
\end{table}

MaSC's PF score is derived at no additional vision-encoder cost from the CP forward pass, yet it outperforms CLIP-T on the same backbone by $\Delta\alpha = 0.027$. This improvement stems entirely from the structural intervention rather than the encoder: the same-backbone control (applying a SigLIP2 global pool with the unstripped prompt) achieves $\alpha = +0.326$, essentially tied with CLIP-T's $+0.327$. Swapping the backbone from CLIP to SigLIP2 yields no meaningful improvement for PF. While MaSC does not establish a new state-of-the-art for PF, its primary contribution is architectural efficiency. Currently, VQAScore ($+0.504$) and ImageReward ($+0.441$) lead the non-LLM lineup, outperforming MaSC by margins of $\Delta\alpha = 0.150$ and $0.087$, respectively. For a deployment already running SigLIP2 for CP, MaSC provides a PF score superior to CLIP-T without an additional vision-encoder forward pass. Furthermore, the foreground-pooled control on the same backbone drops to $\alpha = -0.108$ (Table~\ref{tab:pf_2x3_ablation}). Pooling strictly \emph{onto} the subject inverts the signal, validating the background-masking intervention and confirming that information isolated within the foreground actively degrades PF evaluation.

\begin{table}[t]
    \centering
    \caption{Prompt Following ablation on the SigLIP2~SO400M-NaFlex backbone: image-side pooling ($\{$full, BG, FG$\}$) $\times$ prompt-side subject stripping. Pooled Krippendorff $\alpha$ (Kd$_o$) against pooled DB++ human PF on the shared $6{,}607$-key subset (style excluded). The \textbf{BG-pool $\times$ strip} cell is \textbf{MaSC} (Tab.~\ref{tab:pf_dbplus}). The \textbf{FG-pool} control aligns with Sec.~\ref{sec:results:pf}: pooling \emph{onto} the subject collapses the signal ($\alpha < 0$). \textbf{FG-pool $\times$ subject-stripped} is omitted (---) as evaluating isolated subjects against background-only text is structurally invalid.}
    \label{tab:pf_2x3_ablation}
    \scalebox{0.8}{
    \begin{tabular}{lrr}
        \toprule
        Pool $\backslash$ Prompt & full prompt $\alpha$ & subject-stripped $\alpha$ \\
        \midrule
        BG-pool \,(\textbf{MaSC}) & +0.344 & \textbf{+0.354} \\
        Full pool (no mask) & +0.326 & +0.348 \\
        FG-pool (inverse-mask control) & $-$0.108 & --- \\
        \bottomrule
    \end{tabular}
    }
\end{table}

\subsection{Aggregator Dominates Features}
\label{sec:results:aggregator}

Holding the feature representation fixed, substituting our masked-maxcos aggregator with a mutual-NN foreground-recall aggregator (fraction of mutual NN matches that stay foreground-to-foreground) reduces $\alpha$ by $0.462$ on SigLIP2 patch features and by $0.716$ on DINOv3 patch features. As established in Section~\ref{sec:method:cp}, the strict one-to-one constraint of mutual-NN severely over-penalizes the partial-preservation regime typical of generated outputs. By contrast, the performance gap between SigLIP2 and DINOv3 \emph{under a fixed aggregator} is comparatively minor ($\Delta\alpha = 0.126$).

We also wired the same masked-maxcos aggregator into four specialized correspondence matchers (LoFTR~\cite{sun2021loftr}, RoMa~\cite{Edstedt_2024roma}, MASt3R~\cite{leroy2024groundingimagematching3dmast3r}, SuperPoint+LightGlue~\cite{lindenberger2023lightgluelocalfeaturematching}); all four land $\alpha$-negative on DB++. This demonstrates that the aggregator cannot succeed in isolation; it requires underlying features that natively capture concept identity.

\subsection{Compute and Runtime}
\label{sec:results:runtime}

Table~\ref{tab:runtime} reports per-pair inference latency on a single NVIDIA GeForce RTX 3090 (24GB, driver 550.54.15, CUDA 12.4) paired with an AMD EPYC 7452 (32-core, 128 logical CPUs), running Ubuntu 22.04 with PyTorch 2.6.0 and cuDNN 9.1.0. Timings use fp32, bf16 where the canonical recipe uses it, are taken as the median over 20 timed calls after 5 warmups, and exclude image and mask IO. Mask extraction runs once per image and is cached, so it is not on the per-pair scoring path.

\begin{table}[t]
    \centering
    \caption{Inference cost. Per-pair latency on a single RTX 3090, sorted by latency. Single fixed DB++ pair (\texttt{object\_00\_motorcycle}, \texttt{dreambooth\_sd} seed 0, prompt 0, $512 \times 512$ inputs). Each metric is timed in its own subprocess with \texttt{torch.cuda.synchronize()} around every timed call. \textbf{Bold}: ours.}
    \label{tab:runtime}
    \scalebox{0.8}{
    \begin{tabular}{lcrr}
        \toprule
        Metric & Table & Params (M) & Latency (ms / pair) \\
        \midrule
        CLIP-T                              & PF & 151    & 17.0 \\
        DINO-I                              & CP & 22     & 31.8 \\
        ImageReward                         & PF & 447    & 35.9 \\
        CLIP-I                              & CP & 151    & 36.5 \\
        DINOv3 ViT-L/16 CLS cosine          & CP & 303    & 47.3 \\
        DreamSim                            & CP & 266    & 60.0 \\
        AM-RADIO C-RADIOv4-SO400M           & CP & 431    & 71.8 \\
        SigLIP2 SO400M-NaFlex global pool   & PF & 1\,136 & 89.4 \\
        \textbf{MaSC}                       & PF & 1\,136 & \textbf{91.3} \\
        \textbf{MaSC}                       & CP & 428    & \textbf{160.7} \\
        SigLIP2 SO400M-NaFlex global pool   & CP & 1\,136 & 161.4 \\
        VQAScore                            & PF & 11\,460 & 264.7 \\
        HPSv3                               & PF & 8\,292 & 300.9 \\
        DIFT-SDXL                           & CP & 2\,651 & 1\,399.7 \\
        GPT-4o (API)                        & --- & n/a   & --- \\
        \bottomrule
    \end{tabular}
    }
\end{table}

MaSC's CP path requires $161$ ms per pair, while PF requires $91$ ms. A combined CP and PF deployment requires approximately $190$--$200$ ms per pair; because reference features are cached across all prompts and seeds, the amortized cost is even lower. The lightweight encoder baselines (DINO-I, CLIP-I, DINOv3) are 3--5$\times$ faster but underperform by margins of $\Delta\alpha \geq 0.126$ on DreamBench++ CP. AM-RADIO, at $72$ ms, is the fastest peer at a comparable parameter budget, yet it underperforms by $\Delta\alpha = 0.245$ on DreamBench++ and $\Delta\,\mathrm{AUC} = 0.031$ on ORIDa. DIFT-SDXL is the slowest metric evaluated, requiring $1{,}400$ ms ($\sim$$9\times$ MaSC) while achieving only $\alpha = +0.324$. This suggests that relying on a diffusion backbone introduces substantial computational cost without corresponding performance gains in this setting. The $0.102\,\alpha$ and $0.038\,\mathrm{AUC}$ improvements from the masked-maxcos aggregator introduce zero additional latency: both run at $\sim$$161$ ms because the dominant cost is the shared SigLIP2 forward pass. On the PF side, although CLIP-T ($17$ ms) and ImageReward ($36$ ms) are faster, MaSC's PF ($91$ ms) directly reuses the CP vision features. Compared to other high-performing PF metrics, MaSC is significantly more efficient; VQAScore ($265$ ms) and HPSv3 ($301$ ms) are approximately $3\times$ slower than MaSC while requiring significantly larger parameter budgets.

%% file: src/discussion.tex
\section{Discussion}
\label{sec:discussion}

The main finding is that personalization evaluation benefits more from spatially correct aggregation than from another unmasked global embedding: global image similarity mixes subject identity and scene content, whereas concept preservation and prompt following require different evidence. MaSC enforces this distinction by measuring identity from foreground reference patches with masked-maxcos and measuring scene adherence from a background-only embedding compared with a subject-stripped prompt, which explains why the same SigLIP2 backbone improves when global pooling is replaced by region-conditioned aggregation. The CP results support foreground-local evidence as the proper unit for subject fidelity: MaSC reaches $\alpha=+0.471$ on DreamBench++, outperforming all tested non-LLM baselines and GPT-4V while remaining only $\Delta\alpha=0.028$ below GPT-4o, and reaches $\mathrm{AUC}=0.992$ on ORIDa, outperforming GPT-4o under the evaluated real-photo discrimination protocol. Same-backbone ablations show that the gain comes from masked aggregation rather than encoder choice alone, with improvements of $\Delta\alpha=+0.102$ on DreamBench++ and $\Delta\mathrm{AUC}=+0.038$ on ORIDa, although the failed correspondence-matcher controls show that the aggregator still requires identity-aware features. The PF branch should be read more narrowly: background pooling with subject stripping improves over CLIP-T while reusing the CP visual features, but it does not surpass dedicated reward models such as VQAScore or ImageReward; its contribution is efficiency and diagnostic separation rather than PF state of the art. MaSC also has practical limits: it depends on externally supplied masks (though Appendix~\ref{sec:appendix_sensitivity} demonstrates that performance remains highly robust across different modern segmentation architectures), mask failures affect both filtering and scores, near-empty-mask filtering changes the evaluated population, subject-name stripping is unsuitable for some style prompts, ORIDa lacks human ratings, and the method is designed for single-concept personalization rather than general aesthetic, compositional, or preference evaluation.

%% file: src/conclusion.tex
\section{Conclusion}
\label{sec:conclusion}

We introduced MaSC, a masked similarity metric for concept-driven text-to-image evaluation. Rather than relying on global image embeddings that conflate subject identity with scene content, MaSC spatially decomposes the image: foreground-reference patch matching measures concept preservation, while background-only text-image alignment measures prompt following, both from a shared frozen SigLIP2 forward pass. Across DreamBench++ and ORIDa, MaSC shows that aggregation matters as much as, or more than, encoder choice. It achieves the strongest non-LLM concept-preservation performance on DreamBench++, surpassing GPT-4V and approaching GPT-4o, and nearly perfectly separates same-subject from cross-subject pairs on ORIDa. Its prompt-following score also improves over CLIP-T without an additional vision-encoder pass. MaSC’s main limitation is its dependence on externally supplied foreground masks, which can introduce errors when segmentation fails or subjects are small, absent, or ambiguous. Future work should address mask uncertainty, multi-subject settings, and broader personalization regimes. Overall, our results suggest that spatial attribution should be a central design principle for evaluating generative models.

%% file: src/appendix.tex
\subsection{Sensitivity to Segmentation Source}
\label{sec:appendix_sensitivity}

To verify that MaSC's performance is not artificially tied to the specific capabilities of SAM3, we conducted a sensitivity analysis across four distinct zero-shot segmentation pipelines: SAM3, CLIPSeg~\cite{lueddecke22_cvpr_clipseg}, Grounded-SAM2~\cite{ravi2024sam2, ren2024grounded}, and OWLv2~\cite{minderer2023scaling_owl} + SAM2. We evaluated both the Concept Preservation (CP) and Prompt Following (PF) branches on strictly intersected subsets of DreamBench++ where all four extractors successfully produced a valid mask (5,315 keys for CP, and 4,998 style-excluded keys for PF). 

As shown in Table~\ref{tab:mask_sensitivity_cp}, MaSC's CP score performed consistently across all mask sources. The Krippendorff $\alpha$ varied by at most $\Delta\alpha = 0.012$, and Spearman $\rho$ by at most $\Delta\rho = 0.002$. Notably, the lighter and older CLIPSeg architecture marginally outperformed the SAM3 baseline ($\alpha = +0.482$), demonstrating that our spatial decomposition strategy requires only general subject localization rather than pixel-perfect segmentation boundaries.

Similarly, the PF branch (Table~\ref{tab:mask_sensitivity_pf}) remained highly stable. Because the PF branch uses the mask to suppress foreground tokens, this result confirms that slight variations in segmentation boundaries do not meaningfully alter the background-pooled embedding. Together, these results confirm that MaSC's improvement comes from the structural intervention of spatial decomposition, rather than the isolated accuracy of modern segmentation models.

\begin{table}[h]
    \centering
    \caption{\textbf{CP Sensitivity to Segmentation Source.} Evaluated on a strictly intersected 5,315-key subset. The performance gaps ($\Delta\alpha$, $\Delta\rho$) are relative to the SAM3 baseline. Performance remains highly stable, with the lighter CLIPSeg architecture achieving the highest score.}
    \label{tab:mask_sensitivity_cp}
    \scalebox{0.8}{
    \begin{tabular}{lrrrr}
        \toprule
        Mask Source & $\alpha$ & $\Delta\alpha$ & $\rho$ & $\Delta\rho$ \\
        \midrule
        SAM3 (baseline) & +0.470 & --- & +0.627 & --- \\
        CLIPSeg & +0.482 & +0.012 & +0.630 & +0.002 \\
        Grounded-SAM2 & +0.472 & +0.002 & +0.628 & +0.001 \\
        OWLv2 + SAM2 & +0.473 & +0.003 & +0.626 & -0.001 \\
        \bottomrule
    \end{tabular}
    }
\end{table}

\begin{table}[h]
    \centering
    \caption{\textbf{PF Sensitivity to Segmentation Source.} Evaluated on a strictly intersected 4,998-key subset (style prompts excluded). The background-pooling PF metric is highly robust to mask variations, with maximum score fluctuations of just $\Delta\alpha = 0.005$.}
    \label{tab:mask_sensitivity_pf}
    \scalebox{0.8}{
    \begin{tabular}{lrrrr}
        \toprule
        Mask Source & $\alpha$ & $\Delta\alpha$ & $\rho$ & $\Delta\rho$ \\
        \midrule
        SAM3 (baseline) & +0.395 & --- & +0.555 & --- \\
        CLIPSeg & +0.397 & +0.002 & +0.556 & +0.001 \\
        Grounded-SAM2 & +0.391 & -0.004 & +0.552 & -0.003 \\
        OWLv2 + SAM2 & +0.390 & -0.005 & +0.552 & -0.004 \\
        \bottomrule
    \end{tabular}
    }
\end{table}

\subsection*{N. Licenses, Access Terms, and Attribution}

\begin{table}[h!]
    \centering
    \caption{Dataset licenses and access terms. License information should be
    checked against the official dataset source before public release of any
    derived benchmark tables or redistributed files.}
    \label{tab:eval_card_dataset_licenses}
    \begin{tabular}{p{2.5cm} p{3.0cm} p{6.4cm}}
        \toprule
        Dataset & License / terms & Notes for MaSC use \\
        \midrule
        DreamBench++ & Code: Apache 2.0 (\url{https://github.com/yuangpeng/dreambench\_plus}); image and human rating data: released via Google Drive accompanying the paper, no explicit license stated & The $7{,}135$-key CP subset and $6{,}607$-key PF subset are derived from the DreamBench++ release. We use the image and rating data solely for academic evaluation, consistent with the paper's stated purpose; no subject images or human ratings are redistributed by MaSC, only processed evaluation tables and per-key metric scores. \\
        ORIDa & Dataset access via the authors' download form (Yonsei University); no explicit dataset license is published on the project page. Paper text: CC BY-NC-SA 4.0. & The 50-subject $\times$ 10-environment subset and the corresponding within-/cross-subject pair lists are constructed from ORIDa-train. Provided per-object segmentation masks are used as-is. We treat use as non-commercial academic research, consistent with the paper's CC BY-NC-SA 4.0 release; no images, masks, or derivatives are redistributed by MaSC. The dataset is cited per the authors' attribution requirement, and the exact access terms recorded at download time are preserved in the supplementary manifest. \\
        \bottomrule
    \end{tabular}
\end{table}

\textbf{Dataset redistribution.} The MaSC supplement does not redistribute third-party dataset images, masks, or human ratings unless explicitly allowed by the corresponding license. Instead, it provides scripts and metadata that reproduce the reported scores after the user obtains each dataset under its original terms.

\renewcommand{\arraystretch}{0.9}
\begin{table*}[h]
    \centering
    \footnotesize
    \caption{External model and metric assets used for MaSC and baselines. The exact implementation and checkpoint source are recorded in the supplementary manifest.}
    \label{tab:eval_card_model_licenses}
    \scalebox{0.9}{
    \begin{tabular}{p{2.8cm} p{3.6cm} p{7.2cm}}
        \toprule
        Asset & License / terms & Notes for MaSC use \\
        \midrule
        SigLIP2 SO400M-NaFlex~\cite{tschannen2025siglip} & Apache 2.0 for the evaluated Google/Hugging Face release & Frozen vision--language backbone for MaSC. The exact checkpoint identifier is recorded in the manifest, and license notices are preserved. \\
        SAM3~\cite{carion2025sam3segmentconcepts} & SAM License (Meta); non-exclusive, royalty-free research and commercial use with attribution; prohibited end uses include military or warfare, weapons, ITAR-controlled, nuclear, espionage, and reverse engineering & Concept-mask source on DreamBench++. The exact SAM3 checkpoint is recorded in the manifest, MaSC's use is consistent with the SAM License, and the use of SAM Materials is acknowledged in the paper as required by the license. \\
        CLIP~\cite{radford2021learningtransferablevisualmodels} & MIT for the OpenAI repository code & Used as the CLIP-I appearance baseline and CLIP-T text--image baseline. The exact checkpoint/source is recorded in the manifest, and the MIT license notice is preserved. \\
        DINO~\cite{9709990} & Apache 2.0 (Meta) & Used as the DINO-I baseline via the \texttt{dino\_vits8} checkpoint shipped with DreamBench++. Copyright and license notices are preserved. \\
        DINOv3~\cite{simeoni2025dinov3} & Meta DINOv3 license & Used as the modern self-supervised CP baseline. Use, redistribution, and derivatives must follow Meta's DINOv3 license terms. \\
        DreamSim~\cite{fu2023dreamsim} & MIT (\url{https://github.com/ssundaram21/dreamsim}) & NIGHTS-finetuned perceptual ensemble used as a CP baseline. License notices are preserved. \\
        AM-RADIO C-RADIOv4-SO400M~\cite{Ranzinger_2024_CVPR} & NVIDIA Open Model License Agreement (June 2024); permits commercial and non-commercial use with attribution & Distillation-trained ViT used as a matched-parameter-budget CP baseline. The exact checkpoint and license document are recorded in the manifest; the NVIDIA Open Model License terms are respected. (Note: the related \texttt{RADIO} and \texttt{E-RADIO} variants ship under the NVIDIA Source Code License-NC and are not used in this work.) \\
        DIFT~\cite{tang2023emergent} (Stable Diffusion XL~\cite{rombach2022highresolutionimagesynthesislatent}) & DIFT code: MIT; SDXL: CreativeML Open RAIL++-M & DIFT-SDXL canonical-recipe baseline (timestep $261$, \texttt{up\_ft\_index=1}, null prompt). SDXL weights are used under Stability AI's RAIL++-M terms; DIFT code license notices are preserved. \\
        ImageReward~\cite{xu2023imagereward} & Apache 2.0 (\url{https://github.com/zai-org/ImageReward}) & PF baseline. License notices are preserved. \\
        VQAScore (\texttt{clip-flant5-xxl})~\cite{lin2024evaluating} & Apache 2.0 (\url{https://github.com/linzhiqiu/t2v_metrics}) & PF baseline. License notices are preserved. \\
        HPSv3 (Qwen2-VL-7B-Instruct)~\cite{ma2025hpsv3widespectrumhumanpreference,wang2024qwen2vlenhancingvisionlanguagemodels} & HPSv3 code: MIT; Qwen2-VL-7B-Instruct: Apache 2.0 & PF baseline. The Qwen2-VL-7B-Instruct backbone is used under its Apache 2.0 license; HPSv3 code license notices are preserved. (The 72B variant of Qwen2-VL ships under the Tongyi Qianwen license and is not used in this work.) \\
        LoFTR~\cite{sun2021loftr} & Apache 2.0 & Used as a correspondence-matcher control in the aggregator analysis. Copyright and license notices are preserved. \\
        RoMa~\cite{Edstedt_2024roma} & MIT (\url{https://github.com/Parskatt/RoMa}); the bundled DINOv2 sub-component is Apache 2.0 & Used as a correspondence-matcher control. License notices for both the MIT-licensed RoMa code and the Apache 2.0-licensed DINOv2 sub-component are preserved. \\
        MASt3R~\cite{leroy2024groundingimagematching3dmast3r} & CC BY-NC-SA 4.0; non-commercial use only & Used as a correspondence-matcher control. Non-commercial and ShareAlike restrictions are stated; commercial use may require separate permission. \\
        LightGlue~\cite{lindenberger2023lightgluelocalfeaturematching} & Apache 2.0 & Used as part of the SuperPoint+LightGlue control. Code and pretrained weights are released under Apache 2.0. \\
        SuperPoint~\cite{detone2018superpointselfsupervisedpointdetection} & MIT (\url{https://github.com/rpautrat/SuperPoint}) & Used as part of the SuperPoint+LightGlue control. The open MIT re-implementation by rpautrat is used (not the Magic Leap restrictive release); license notices are preserved. \\
        GPT-4V / GPT-4o~\cite{openai2024gpt4technicalreport,openai2024gpt4ocard} & OpenAI API; OpenAI Terms of Use & LLM-judge baselines. DreamBench++ scores are reused from the public release; ORIDa scores are obtained via the OpenAI API in accordance with OpenAI's terms. \\
        SAM 2~\cite{ravi2024sam2} & Apache 2.0 (Meta) for code and model checkpoints; the bundled \texttt{cc\_torch} post-processing utility is BSD-3-Clause & Concept-mask source / segmentation control. The exact SAM 2 checkpoint is recorded in the manifest, and Apache 2.0 license notices are preserved. \\
        Grounded SAM~\cite{ren2024grounded} & Apache 2.0 (\url{https://github.com/IDEA-Research/Grounded-Segment-Anything}); bundles Grounding DINO (Apache 2.0) and Segment Anything (see SAM/SAM 2 rows) & Open-vocabulary segmentation control. The Grounded SAM glue code is used under Apache 2.0; bundled component licenses are honored independently and recorded in the manifest. \\
        CLIPSeg~\cite{lueddecke22_cvpr_clipseg} & Source code: MIT (\url{https://github.com/timojl/clipseg}); released weights (\texttt{CIDAS/clipseg-rd64-refined} on Hugging Face): Apache 2.0 & Text-prompted segmentation control. The exact checkpoint is recorded in the manifest; both the MIT code license and the Apache 2.0 weights license are preserved. \\
        OWLv2~\cite{minderer2023scaling_owl} & Apache 2.0 (Google) for both code and model checkpoints & Open-vocabulary detection control. The exact checkpoint identifier is recorded in the manifest, and Apache 2.0 license notices are preserved. \\
        \bottomrule
    \end{tabular}
    }
\end{table*}

%% file: checklist.tex
\section*{NeurIPS Paper Checklist}

\begin{enumerate}

\item {\bf Claims}
    \item[] Question: Do the main claims made in the abstract and introduction accurately reflect the paper's contributions and scope?
    \item[] Answer: \answerYes{} % Replace by \answerYes{}, \answerNo{}, or \answerNA{}.
    \item[] Justification: The abstract and introduction state that MaSC is a unified concept-preservation and prompt-following metric that scores both signals from a single SigLIP2 forward pass given an externally supplied concept mask. The five claimed contributions — DreamBench++ CP performance, ORIDa identity discrimination, the same-backbone ablation, the matched-budget comparison, and the PF result — each correspond to a specific empirical evaluation reported in the paper. The scope is single-concept text-to-image personalization, and the paper explicitly does not claim PF state-of-the-art.

\item {\bf Limitations}
    \item[] Question: Does the paper discuss the limitations of the work performed by the authors?
    \item[] Answer: \answerYes{} % Replace by \answerYes{}, \answerNo{}, or \answerNA{}.
    \item[] Justification: The paper discusses its limitations in the Discussion and Conclusion: dependence on externally supplied foreground masks, the effect of mask-failure filtering on the evaluated population, the inapplicability of subject-name stripping to style prompts, the absence of human ratings on ORIDa, and the restriction to single-concept personalization. The Prompt Following section explicitly notes that MaSC does not establish PF state-of-the-art and is positioned as a free byproduct of the concept-preservation forward pass rather than a dedicated reward model.

\item {\bf Theory assumptions and proofs}
    \item[] Question: For each theoretical result, does the paper provide the full set of assumptions and a complete (and correct) proof?
    \item[] Answer: \answerNA{} % Replace by \answerYes{}, \answerNo{}, or \answerNA{}.
    \item[] Justification: The paper does not present formal theoretical results, theorems, or proofs. The mathematical content defines the masked-maxcos concept-preservation score, the background-pooled prompt-following score, and the subject-stripping operator used in the empirical evaluation.

    \item {\bf Experimental result reproducibility}
    \item[] Question: Does the paper fully disclose all the information needed to reproduce the main experimental results of the paper to the extent that it affects the main claims and/or conclusions of the paper (regardless of whether the code and data are provided or not)?
    \item[] Answer: \answerYes{} % Replace by \answerYes{}, \answerNo{}, or \answerNA{}.
    \item[] Justification: The paper specifies the metric definitions, the frozen public SigLIP2 SO400M-NaFlex backbone, the mask source and filtering protocol, and the dataset-specific subject sampling for ORIDa. Comparator configurations — including the DIFT-SDXL canonical recipe, the DINOv3 and DreamSim variants, AM-RADIO, VQAScore, ImageReward, and HPSv3 — are documented in the related-work and results sections. Both evaluation datasets (DreamBench++ and ORIDa) are publicly available, and the paper announces release of a pip-installable Python package for the metric and comparator reproductions.

\item {\bf Open access to data and code}
    \item[] Question: Does the paper provide open access to the data and code, with sufficient instructions to faithfully reproduce the main experimental results, as described in supplemental material?
    \item[] Answer: \answerYes{} % Replace by \answerYes{}, \answerNo{}, or \answerNA{}.
    \item[] Justification: We provide anonymized supplementary code (https://anonymous.4open.science/r/masc-reproduction-3536) containing evaluation scripts to reproduce the MaSC scores, the same-backbone and aggregator ablations, and the reported correlations and AUCs against every comparator. The metric is also released as a pip-installable Python package (masc-metric). The raw datasets are existing public benchmarks (DreamBench++ and ORIDa); the supplementary material describes how to obtain them and reproduce the processed evaluation tables.

\item {\bf Experimental setting/details}
    \item[] Question: Does the paper specify all the training and test details (e.g., data splits, hyperparameters, how they were chosen, type of optimizer) necessary to understand the results?
    \item[] Answer: \answerYes{} % Replace by \answerYes{}, \answerNo{}, or \answerNA{}.
    \item[] Justification: The paper is training-free; it specifies the frozen backbone, the mask source and filtering thresholds, the DreamBench++ subset sizes for CP and PF, and the ORIDa sampling protocol used for the within- and cross-subject pair construction. Comparator configurations — including the DIFT-SDXL canonical recipe, DINOv3 and DreamSim variants, AM-RADIO, VQAScore, ImageReward, and HPSv3 — are documented alongside the results, and the evaluation statistics are stated for each table.

\item {\bf Experiment statistical significance}
    \item[] Question: Does the paper report error bars suitably and correctly defined or other appropriate information about the statistical significance of the experiments?
    \item[] Answer: \answerYes{} % Replace by \answerYes{}, \answerNo{}, or \answerNA{}.
    \item[] Justification: The metric is deterministic, so there is no run-to-run variance to report. The headline numbers are computed over the full 7{,}135-key DreamBench++ apples-to-apples subset and the 4{,}500 ORIDa pairs; the ORIDa table additionally reports per-pair within- and cross-subject score distributions with standard deviations. A same-backbone ablation isolates the spatial decomposition from the encoder.

\item {\bf Experiments compute resources}
    \item[] Question: For each experiment, does the paper provide sufficient information on the computer resources (type of compute workers, memory, time of execution) needed to reproduce the experiments?
    \item[] Answer: \answerYes{} % Replace by \answerYes{}, \answerNo{}, or \answerNA{}.
    \item[] Justification: The runtime table reports per-pair latency in milliseconds and parameter counts for each evaluated metric. The hardware and software stack used for the benchmark is also specified, including the NVIDIA GeForce RTX 3090 GPU, CUDA version, AMD EPYC 7452 CPU, logical-CPU count, operating system, PyTorch, and cuDNN versions, along with the timing protocol (median over 20 calls after 5 warmups, image and mask IO excluded).
    
\item {\bf Code of ethics}
    \item[] Question: Does the research conducted in the paper conform, in every respect, with the NeurIPS Code of Ethics \url{https://neurips.cc/public/EthicsGuidelines}?
    \item[] Answer: \answerYes{} % Replace by \answerYes{}, \answerNo{}, or \answerNA{}.
    \item[] Justification: The research uses existing public personalization benchmarks and pretrained models for evaluating text-to-image personalization metrics, and does not involve human-subject experiments, private data collection, or deployment decisions. We have reviewed the NeurIPS Code of Ethics and believe the work conforms to it.

\item {\bf Broader impacts}
    \item[] Question: Does the paper discuss both potential positive societal impacts and negative societal impacts of the work performed?
    \item[] Answer: \answerYes{} % Replace by \answerYes{}, \answerNo{}, or \answerNA{}.
    \item[] Justification: The positive impact of MaSC is that it provides a fast, reproducible, non-API personalization metric, lowering the compute and cost barrier to rigorous benchmarking of personalization methods and reducing reliance on opaque LLM judges. Potential negative impacts include over-reliance on a proxy metric in deployment-critical settings, or its use to accelerate optimization of subject-driven generation models that can be misused for non-consensual or deceptive imagery; MaSC should therefore be used as a diagnostic evaluation tool rather than as a substitute for human review in sensitive applications.
    
\item {\bf Safeguards}
    \item[] Question: Does the paper describe safeguards that have been put in place for responsible release of data or models that have a high risk for misuse (e.g., pre-trained language models, image generators, or scraped datasets)?
    \item[] Answer: \answerNA{} % Replace by \answerYes{}, \answerNo{}, or \answerNA{}.
    \item[] Justification: The paper does not release high-risk generative models, pretrained language models, scraped datasets, or data intended for direct deployment in sensitive applications. The released assets are evaluation code, metric scripts, and a pip-installable package wrapping a frozen public SigLIP2 checkpoint.

\item {\bf Licenses for existing assets}
    \item[] Question: Are the creators or original owners of assets (e.g., code, data, models), used in the paper, properly credited and are the license and terms of use explicitly mentioned and properly respected?
    \item[] Answer: \answerYes{} % Replace by \answerYes{}, \answerNo{}, or \answerNA{}.
    \item[] Justification: The paper cites the original sources for all datasets, pretrained models, and metric implementations used in the evaluation. The supplementary material lists the license or access terms for each asset, including DreamBench++, ORIDa, SigLIP2, SAM3, DINO, DINOv3, DreamSim, AM-RADIO, DIFT/Stable~Diffusion~XL, CLIP, ImageReward, VQAScore, HPSv3, the correspondence-matcher controls (LoFTR, RoMa, MASt3R, LightGlue, SuperPoint), and the GPT-4V/4o API judges. The experiments are conducted in accordance with these terms, and raw third-party datasets, masks, and human ratings are not redistributed.

\item {\bf New assets}
    \item[] Question: Are new assets introduced in the paper well documented and is the documentation provided alongside the assets?
    \item[] Answer: \answerYes{} % Replace by \answerYes{}, \answerNo{}, or \answerNA{}.
    \item[] Justification: The paper introduces the MaSC evaluation code, comparator reproduction scripts, and the pip-installable masc-metric package. These assets are documented in the supplementary material, including expected inputs, mask-source assumptions, filtering thresholds, metric computation, comparator configurations, runtime assumptions, and known limitations.

\item {\bf Crowdsourcing and research with human subjects}
    \item[] Question: For crowdsourcing experiments and research with human subjects, does the paper include the full text of instructions given to participants and screenshots, if applicable, as well as details about compensation (if any)? 
    \item[] Answer: \answerNA{} % Replace by \answerYes{}, \answerNo{}, or \answerNA{}.
    \item[] Justification: The paper does not involve crowdsourcing, user studies, annotation by human participants, or research with human subjects. All evaluations are performed on existing public benchmarks; the human ratings used as ground truth on DreamBench++ are obtained as-is from the original release.

\item {\bf Institutional review board (IRB) approvals or equivalent for research with human subjects}
    \item[] Question: Does the paper describe potential risks incurred by study participants, whether such risks were disclosed to the subjects, and whether Institutional Review Board (IRB) approvals (or an equivalent approval/review based on the requirements of your country or institution) were obtained?
    \item[] Answer: \answerNA{} % Replace by \answerYes{}, \answerNo{}, or \answerNA{}.
    \item[] Justification: The paper does not involve human-subject research, crowdsourcing, collection of personal data, or interaction with study participants. Therefore, IRB or equivalent approval is not applicable.

\item {\bf Declaration of LLM usage}
    \item[] Question: Does the paper describe the usage of LLMs if it is an important, original, or non-standard component of the core methods in this research? Note that if the LLM is used only for writing, editing, or formatting purposes and does \emph{not} impact the core methodology, scientific rigor, or originality of the research, declaration is not required.
    %this research? 
    \item[] Answer: \answerNA{} % Replace by \answerYes{}, \answerNo{}, or \answerNA{}.
    \item[] Justification: LLMs are not used as a component of MaSC, the experiments, or the metric computation. GPT-4V and GPT-4o appear only as comparator baselines, not as part of MaSC itself. Any use of LLMs during the project was limited to writing assistance and code-implementation help (analogous to standard developer tooling) and did not affect the formulation of the metric, the experimental design, the interpretation of results, or the originality of the research.

\end{enumerate}

%% file: main.bib
@inproceedings{peng2024dreambench,
    author={Yuang Peng and Yuxin Cui and Haomiao Tang and Zekun Qi and Runpei Dong and Jing Bai and Chunrui Han and Zheng Ge and Xiangyu Zhang and Shu-Tao Xia},
    title={DreamBench++: A Human-Aligned Benchmark for Personalized Image Generation},
    booktitle={The Thirteenth International Conference on Learning Representations},
    url={https://openreview.net/forum?id=4GSOESJrk6},
    year={2025},
}

@inproceedings{kim2025orida,
    title={ORIDa: Object-centric Real-world Image Composition Dataset},
    author={Kim, Jinwoo and Han, Sangmin and Jeong, Jinho and Choi, Jiwoo and Kim, Dongyeoung and Kim, Seon Joo},
    booktitle={Proceedings of the Computer Vision and Pattern Recognition Conference},
    pages={3051--3060},
    year={2025}
}

@article{krippendorff,
author = {krippendorff, klaus},
year = {2011},
month = {01},
pages = {},
title = {Computing Krippendorff's Alpha-Reliability}
}

@article{tschannen2025siglip,
    title={SigLIP 2: Multilingual Vision-Language Encoders with Improved Semantic Understanding, Localization, and Dense Features},
    author={Tschannen, Michael and Gritsenko, Alexey and Wang, Xiao and Naeem, Muhammad Ferjad and Alabdulmohsin, Ibrahim and Parthasarathy, Nikhil and Evans, Talfan and Beyer, Lucas and Xia, Ye and Mustafa, Basil and H\'enaff, Olivier and Harmsen, Jeremiah and Steiner, Andreas and Zhai, Xiaohua},
    year={2025},
    journal={arXiv preprint arXiv:2502.14786}
}

@misc{carion2025sam3segmentconcepts,
    title={SAM 3: Segment Anything with Concepts},
    author={Nicolas Carion and Laura Gustafson and Yuan-Ting Hu and Shoubhik Debnath and Ronghang Hu and Didac Suris and Chaitanya Ryali and Kalyan Vasudev Alwala and Haitham Khedr and Andrew Huang and Jie Lei and Tengyu Ma and Baishan Guo and Arpit Kalla and Markus Marks and Joseph Greer and Meng Wang and Peize Sun and Roman Rädle and Triantafyllos Afouras and Effrosyni Mavroudi and Katherine Xu and Tsung-Han Wu and Yu Zhou and Liliane Momeni and Rishi Hazra and Shuangrui Ding and Sagar Vaze and Francois Porcher and Feng Li and Siyuan Li and Aishwarya Kamath and Ho Kei Cheng and Piotr Dollár and Nikhila Ravi and Kate Saenko and Pengchuan Zhang and Christoph Feichtenhofer},
    year={2025},
    eprint={2511.16719},
    archivePrefix={arXiv},
    primaryClass={cs.CV},
    url={https://arxiv.org/abs/2511.16719},
}

@inproceedings{fu2023dreamsim,
    title={DreamSim: Learning New Dimensions of Human Visual Similarity using Synthetic Data},
    author= {Fu, Stephanie and Tamir, Netanel and Sundaram, Shobhita and Chai, Lucy and Zhang, Richard and Dekel, Tali and Isola, Phillip},
    booktitle={Advances in Neural Information Processing Systems},
    pages={50742--50768},
    volume={36},
    year={2023}
}

@InProceedings{Ranzinger_2024_CVPR,
    author    = {Ranzinger, Mike and Heinrich, Greg and Kautz, Jan and Molchanov, Pavlo},
    title     = {AM-RADIO: Agglomerative Vision Foundation Model Reduce All Domains Into One},
    booktitle = {Proceedings of the IEEE/CVF Conference on Computer Vision and Pattern Recognition (CVPR)},
    month     = {June},
    year      = {2024},
    pages     = {12490-12500}
}

@misc{simeoni2025dinov3,
    title={{DINOv3}},
    author={Sim{\'e}oni, Oriane and Vo, Huy V. and Seitzer, Maximilian and Baldassarre, Federico and Oquab, Maxime and Jose, Cijo and Khalidov, Vasil and Szafraniec, Marc and Yi, Seungeun and Ramamonjisoa, Micha{\"e}l and Massa, Francisco and Haziza, Daniel and Wehrstedt, Luca and Wang, Jianyuan and Darcet, Timoth{\'e}e and Moutakanni, Th{\'e}o and Sentana, Leonel and Roberts, Claire and Vedaldi, Andrea and Tolan, Jamie and Brandt, John and Couprie, Camille and Mairal, Julien and J{\'e}gou, Herv{\'e} and Labatut, Patrick and Bojanowski, Piotr},
    year={2025},
    eprint={2508.10104},
    archivePrefix={arXiv},
    primaryClass={cs.CV},
    url={https://arxiv.org/abs/2508.10104},
}

@inproceedings{tang2023emergent,
    title={Emergent Correspondence from Image Diffusion},
    author={Luming Tang and Menglin Jia and Qianqian Wang and Cheng Perng Phoo and Bharath Hariharan},
    booktitle={Thirty-seventh Conference on Neural Information Processing Systems},
    year={2023},
    url={https://openreview.net/forum?id=ypOiXjdfnU}
}

@misc{radford2021learningtransferablevisualmodels,
    title={Learning Transferable Visual Models From Natural Language Supervision}, 
    author={Alec Radford and Jong Wook Kim and Chris Hallacy and Aditya Ramesh and Gabriel Goh and Sandhini Agarwal and Girish Sastry and Amanda Askell and Pamela Mishkin and Jack Clark and Gretchen Krueger and Ilya Sutskever},
    year={2021},
    eprint={2103.00020},
    archivePrefix={arXiv},
    primaryClass={cs.CV},
    url={https://arxiv.org/abs/2103.00020}, 
}

@INPROCEEDINGS{9709990,
    author={Caron, Mathilde and Touvron, Hugo and Misra, Ishan and Jegou, Hervé and Mairal, Julien and Bojanowski, Piotr and Joulin, Armand},
    booktitle={2021 IEEE/CVF International Conference on Computer Vision (ICCV)}, 
    title={Emerging Properties in Self-Supervised Vision Transformers}, 
    year={2021},
    volume={},
    number={},
    pages={9630-9640},
    doi={10.1109/ICCV48922.2021.00951}
}

@article{lin2024evaluating,
    title={Evaluating Text-to-Visual Generation with Image-to-Text Generation},
    author={Lin, Zhiqiu and Pathak, Deepak and Li, Baiqi and Li, Jiayao and Xia, Xide and Neubig, Graham and Zhang, Pengchuan and Ramanan, Deva},
    journal={arXiv preprint arXiv:2404.01291},
    year={2024}
}

@inproceedings{xu2023imagereward,
    title={ImageReward: learning and evaluating human preferences for text-to-image generation},
    author={Xu, Jiazheng and Liu, Xiao and Wu, Yuchen and Tong, Yuxuan and Li, Qinkai and Ding, Ming and Tang, Jie and Dong, Yuxiao},
    booktitle={Proceedings of the 37th International Conference on Neural Information Processing Systems},
    pages={15903--15935},
    year={2023}
}

@misc{ma2025hpsv3widespectrumhumanpreference,
    title={HPSv3: Towards Wide-Spectrum Human Preference Score}, 
    author={Yuhang Ma and Xiaoshi Wu and Keqiang Sun and Hongsheng Li},
    year={2025},
    eprint={2508.03789},
    archivePrefix={arXiv},
    primaryClass={cs.CV},
    url={https://arxiv.org/abs/2508.03789}, 
}

@misc{openai2024gpt4ocard,
    title={GPT-4o System Card}, 
    author={OpenAI and : and Aaron Hurst and Adam Lerer and Adam P. Goucher and Adam Perelman and Aditya Ramesh and Aidan Clark and AJ Ostrow and Akila Welihinda and Alan Hayes and Alec Radford and Aleksander Mądry and Alex Baker-Whitcomb and Alex Beutel and Alex Borzunov and Alex Carney and Alex Chow and Alex Kirillov and Alex Nichol and Alex Paino and Alex Renzin and Alex Tachard Passos and Alexander Kirillov and Alexi Christakis and Alexis Conneau and Ali Kamali and Allan Jabri and Allison Moyer and Allison Tam and Amadou Crookes and Amin Tootoochian and Amin Tootoonchian and Ananya Kumar and Andrea Vallone and Andrej Karpathy and Andrew Braunstein and Andrew Cann and Andrew Codispoti and Andrew Galu and Andrew Kondrich and Andrew Tulloch and Andrey Mishchenko and Angela Baek and Angela Jiang and Antoine Pelisse and Antonia Woodford and Anuj Gosalia and Arka Dhar and Ashley Pantuliano and Avi Nayak and Avital Oliver and Barret Zoph and Behrooz Ghorbani and Ben Leimberger and Ben Rossen and Ben Sokolowsky and Ben Wang and Benjamin Zweig and Beth Hoover and Blake Samic and Bob McGrew and Bobby Spero and Bogo Giertler and Bowen Cheng and Brad Lightcap and Brandon Walkin and Brendan Quinn and Brian Guarraci and Brian Hsu and Bright Kellogg and Brydon Eastman and Camillo Lugaresi and Carroll Wainwright and Cary Bassin and Cary Hudson and Casey Chu and Chad Nelson and Chak Li and Chan Jun Shern and Channing Conger and Charlotte Barette and Chelsea Voss and Chen Ding and Cheng Lu and Chong Zhang and Chris Beaumont and Chris Hallacy and Chris Koch and Christian Gibson and Christina Kim and Christine Choi and Christine McLeavey and Christopher Hesse and Claudia Fischer and Clemens Winter and Coley Czarnecki and Colin Jarvis and Colin Wei and Constantin Koumouzelis and Dane Sherburn and Daniel Kappler and Daniel Levin and Daniel Levy and David Carr and David Farhi and David Mely and David Robinson and David Sasaki and Denny Jin and Dev Valladares and Dimitris Tsipras and Doug Li and Duc Phong Nguyen and Duncan Findlay and Edede Oiwoh and Edmund Wong and Ehsan Asdar and Elizabeth Proehl and Elizabeth Yang and Eric Antonow and Eric Kramer and Eric Peterson and Eric Sigler and Eric Wallace and Eugene Brevdo and Evan Mays and Farzad Khorasani and Felipe Petroski Such and Filippo Raso and Francis Zhang and Fred von Lohmann and Freddie Sulit and Gabriel Goh and Gene Oden and Geoff Salmon and Giulio Starace and Greg Brockman and Hadi Salman and Haiming Bao and Haitang Hu and Hannah Wong and Haoyu Wang and Heather Schmidt and Heather Whitney and Heewoo Jun and Hendrik Kirchner and Henrique Ponde de Oliveira Pinto and Hongyu Ren and Huiwen Chang and Hyung Won Chung and Ian Kivlichan and Ian O'Connell and Ian O'Connell and Ian Osband and Ian Silber and Ian Sohl and Ibrahim Okuyucu and Ikai Lan and Ilya Kostrikov and Ilya Sutskever and Ingmar Kanitscheider and Ishaan Gulrajani and Jacob Coxon and Jacob Menick and Jakub Pachocki and James Aung and James Betker and James Crooks and James Lennon and Jamie Kiros and Jan Leike and Jane Park and Jason Kwon and Jason Phang and Jason Teplitz and Jason Wei and Jason Wolfe and Jay Chen and Jeff Harris and Jenia Varavva and Jessica Gan Lee and Jessica Shieh and Ji Lin and Jiahui Yu and Jiayi Weng and Jie Tang and Jieqi Yu and Joanne Jang and Joaquin Quinonero Candela and Joe Beutler and Joe Landers and Joel Parish and Johannes Heidecke and John Schulman and Jonathan Lachman and Jonathan McKay and Jonathan Uesato and Jonathan Ward and Jong Wook Kim and Joost Huizinga and Jordan Sitkin and Jos Kraaijeveld and Josh Gross and Josh Kaplan and Josh Snyder and Joshua Achiam and Joy Jiao and Joyce Lee and Juntang Zhuang and Justyn Harriman and Kai Fricke and Kai Hayashi and Karan Singhal and Katy Shi and Kavin Karthik and Kayla Wood and Kendra Rimbach and Kenny Hsu and Kenny Nguyen and Keren Gu-Lemberg and Kevin Button and Kevin Liu and Kiel Howe and Krithika Muthukumar and Kyle Luther and Lama Ahmad and Larry Kai and Lauren Itow and Lauren Workman and Leher Pathak and Leo Chen and Li Jing and Lia Guy and Liam Fedus and Liang Zhou and Lien Mamitsuka and Lilian Weng and Lindsay McCallum and Lindsey Held and Long Ouyang and Louis Feuvrier and Lu Zhang and Lukas Kondraciuk and Lukasz Kaiser and Luke Hewitt and Luke Metz and Lyric Doshi and Mada Aflak and Maddie Simens and Madelaine Boyd and Madeleine Thompson and Marat Dukhan and Mark Chen and Mark Gray and Mark Hudnall and Marvin Zhang and Marwan Aljubeh and Mateusz Litwin and Matthew Zeng and Max Johnson and Maya Shetty and Mayank Gupta and Meghan Shah and Mehmet Yatbaz and Meng Jia Yang and Mengchao Zhong and Mia Glaese and Mianna Chen and Michael Janner and Michael Lampe and Michael Petrov and Michael Wu and Michele Wang and Michelle Fradin and Michelle Pokrass and Miguel Castro and Miguel Oom Temudo de Castro and Mikhail Pavlov and Miles Brundage and Miles Wang and Minal Khan and Mira Murati and Mo Bavarian and Molly Lin and Murat Yesildal and Nacho Soto and Natalia Gimelshein and Natalie Cone and Natalie Staudacher and Natalie Summers and Natan LaFontaine and Neil Chowdhury and Nick Ryder and Nick Stathas and Nick Turley and Nik Tezak and Niko Felix and Nithanth Kudige and Nitish Keskar and Noah Deutsch and Noel Bundick and Nora Puckett and Ofir Nachum and Ola Okelola and Oleg Boiko and Oleg Murk and Oliver Jaffe and Olivia Watkins and Olivier Godement and Owen Campbell-Moore and Patrick Chao and Paul McMillan and Pavel Belov and Peng Su and Peter Bak and Peter Bakkum and Peter Deng and Peter Dolan and Peter Hoeschele and Peter Welinder and Phil Tillet and Philip Pronin and Philippe Tillet and Prafulla Dhariwal and Qiming Yuan and Rachel Dias and Rachel Lim and Rahul Arora and Rajan Troll and Randall Lin and Rapha Gontijo Lopes and Raul Puri and Reah Miyara and Reimar Leike and Renaud Gaubert and Reza Zamani and Ricky Wang and Rob Donnelly and Rob Honsby and Rocky Smith and Rohan Sahai and Rohit Ramchandani and Romain Huet and Rory Carmichael and Rowan Zellers and Roy Chen and Ruby Chen and Ruslan Nigmatullin and Ryan Cheu and Saachi Jain and Sam Altman and Sam Schoenholz and Sam Toizer and Samuel Miserendino and Sandhini Agarwal and Sara Culver and Scott Ethersmith and Scott Gray and Sean Grove and Sean Metzger and Shamez Hermani and Shantanu Jain and Shengjia Zhao and Sherwin Wu and Shino Jomoto and Shirong Wu and Shuaiqi and Xia and Sonia Phene and Spencer Papay and Srinivas Narayanan and Steve Coffey and Steve Lee and Stewart Hall and Suchir Balaji and Tal Broda and Tal Stramer and Tao Xu and Tarun Gogineni and Taya Christianson and Ted Sanders and Tejal Patwardhan and Thomas Cunninghman and Thomas Degry and Thomas Dimson and Thomas Raoux and Thomas Shadwell and Tianhao Zheng and Todd Underwood and Todor Markov and Toki Sherbakov and Tom Rubin and Tom Stasi and Tomer Kaftan and Tristan Heywood and Troy Peterson and Tyce Walters and Tyna Eloundou and Valerie Qi and Veit Moeller and Vinnie Monaco and Vishal Kuo and Vlad Fomenko and Wayne Chang and Weiyi Zheng and Wenda Zhou and Wesam Manassra and Will Sheu and Wojciech Zaremba and Yash Patil and Yilei Qian and Yongjik Kim and Youlong Cheng and Yu Zhang and Yuchen He and Yuchen Zhang and Yujia Jin and Yunxing Dai and Yury Malkov},
    year={2024},
    eprint={2410.21276},
    archivePrefix={arXiv},
    primaryClass={cs.CL},
    url={https://arxiv.org/abs/2410.21276}, 
}

@misc{openai2024gpt4technicalreport,
      title={GPT-4 Technical Report}, 
      author={OpenAI and Josh Achiam and Steven Adler and Sandhini Agarwal and Lama Ahmad and Ilge Akkaya and Florencia Leoni Aleman and Diogo Almeida and Janko Altenschmidt and Sam Altman and Shyamal Anadkat and Red Avila and Igor Babuschkin and Suchir Balaji and Valerie Balcom and Paul Baltescu and Haiming Bao and Mohammad Bavarian and Jeff Belgum and Irwan Bello and Jake Berdine and Gabriel Bernadett-Shapiro and Christopher Berner and Lenny Bogdonoff and Oleg Boiko and Madelaine Boyd and Anna-Luisa Brakman and Greg Brockman and Tim Brooks and Miles Brundage and Kevin Button and Trevor Cai and Rosie Campbell and Andrew Cann and Brittany Carey and Chelsea Carlson and Rory Carmichael and Brooke Chan and Che Chang and Fotis Chantzis and Derek Chen and Sully Chen and Ruby Chen and Jason Chen and Mark Chen and Ben Chess and Chester Cho and Casey Chu and Hyung Won Chung and Dave Cummings and Jeremiah Currier and Yunxing Dai and Cory Decareaux and Thomas Degry and Noah Deutsch and Damien Deville and Arka Dhar and David Dohan and Steve Dowling and Sheila Dunning and Adrien Ecoffet and Atty Eleti and Tyna Eloundou and David Farhi and Liam Fedus and Niko Felix and Simón Posada Fishman and Juston Forte and Isabella Fulford and Leo Gao and Elie Georges and Christian Gibson and Vik Goel and Tarun Gogineni and Gabriel Goh and Rapha Gontijo-Lopes and Jonathan Gordon and Morgan Grafstein and Scott Gray and Ryan Greene and Joshua Gross and Shixiang Shane Gu and Yufei Guo and Chris Hallacy and Jesse Han and Jeff Harris and Yuchen He and Mike Heaton and Johannes Heidecke and Chris Hesse and Alan Hickey and Wade Hickey and Peter Hoeschele and Brandon Houghton and Kenny Hsu and Shengli Hu and Xin Hu and Joost Huizinga and Shantanu Jain and Shawn Jain and Joanne Jang and Angela Jiang and Roger Jiang and Haozhun Jin and Denny Jin and Shino Jomoto and Billie Jonn and Heewoo Jun and Tomer Kaftan and Łukasz Kaiser and Ali Kamali and Ingmar Kanitscheider and Nitish Shirish Keskar and Tabarak Khan and Logan Kilpatrick and Jong Wook Kim and Christina Kim and Yongjik Kim and Jan Hendrik Kirchner and Jamie Kiros and Matt Knight and Daniel Kokotajlo and Łukasz Kondraciuk and Andrew Kondrich and Aris Konstantinidis and Kyle Kosic and Gretchen Krueger and Vishal Kuo and Michael Lampe and Ikai Lan and Teddy Lee and Jan Leike and Jade Leung and Daniel Levy and Chak Ming Li and Rachel Lim and Molly Lin and Stephanie Lin and Mateusz Litwin and Theresa Lopez and Ryan Lowe and Patricia Lue and Anna Makanju and Kim Malfacini and Sam Manning and Todor Markov and Yaniv Markovski and Bianca Martin and Katie Mayer and Andrew Mayne and Bob McGrew and Scott Mayer McKinney and Christine McLeavey and Paul McMillan and Jake McNeil and David Medina and Aalok Mehta and Jacob Menick and Luke Metz and Andrey Mishchenko and Pamela Mishkin and Vinnie Monaco and Evan Morikawa and Daniel Mossing and Tong Mu and Mira Murati and Oleg Murk and David Mély and Ashvin Nair and Reiichiro Nakano and Rajeev Nayak and Arvind Neelakantan and Richard Ngo and Hyeonwoo Noh and Long Ouyang and Cullen O'Keefe and Jakub Pachocki and Alex Paino and Joe Palermo and Ashley Pantuliano and Giambattista Parascandolo and Joel Parish and Emy Parparita and Alex Passos and Mikhail Pavlov and Andrew Peng and Adam Perelman and Filipe de Avila Belbute Peres and Michael Petrov and Henrique Ponde de Oliveira Pinto and Michael and Pokorny and Michelle Pokrass and Vitchyr H. Pong and Tolly Powell and Alethea Power and Boris Power and Elizabeth Proehl and Raul Puri and Alec Radford and Jack Rae and Aditya Ramesh and Cameron Raymond and Francis Real and Kendra Rimbach and Carl Ross and Bob Rotsted and Henri Roussez and Nick Ryder and Mario Saltarelli and Ted Sanders and Shibani Santurkar and Girish Sastry and Heather Schmidt and David Schnurr and John Schulman and Daniel Selsam and Kyla Sheppard and Toki Sherbakov and Jessica Shieh and Sarah Shoker and Pranav Shyam and Szymon Sidor and Eric Sigler and Maddie Simens and Jordan Sitkin and Katarina Slama and Ian Sohl and Benjamin Sokolowsky and Yang Song and Natalie Staudacher and Felipe Petroski Such and Natalie Summers and Ilya Sutskever and Jie Tang and Nikolas Tezak and Madeleine B. Thompson and Phil Tillet and Amin Tootoonchian and Elizabeth Tseng and Preston Tuggle and Nick Turley and Jerry Tworek and Juan Felipe Cerón Uribe and Andrea Vallone and Arun Vijayvergiya and Chelsea Voss and Carroll Wainwright and Justin Jay Wang and Alvin Wang and Ben Wang and Jonathan Ward and Jason Wei and CJ Weinmann and Akila Welihinda and Peter Welinder and Jiayi Weng and Lilian Weng and Matt Wiethoff and Dave Willner and Clemens Winter and Samuel Wolrich and Hannah Wong and Lauren Workman and Sherwin Wu and Jeff Wu and Michael Wu and Kai Xiao and Tao Xu and Sarah Yoo and Kevin Yu and Qiming Yuan and Wojciech Zaremba and Rowan Zellers and Chong Zhang and Marvin Zhang and Shengjia Zhao and Tianhao Zheng and Juntang Zhuang and William Zhuk and Barret Zoph},
      year={2024},
      eprint={2303.08774},
      archivePrefix={arXiv},
      primaryClass={cs.CL},
      url={https://arxiv.org/abs/2303.08774}, 
}

@misc{gal2022imageworthwordpersonalizing,
    title={An Image is Worth One Word: Personalizing Text-to-Image Generation using Textual Inversion}, 
    author={Rinon Gal and Yuval Alaluf and Yuval Atzmon and Or Patashnik and Amit H. Bermano and Gal Chechik and Daniel Cohen-Or},
    year={2022},
    eprint={2208.01618},
    archivePrefix={arXiv},
    primaryClass={cs.CV},
    url={https://arxiv.org/abs/2208.01618}, 
}

@article{ruiz2022dreambooth,
    title={DreamBooth: Fine Tuning Text-to-image Diffusion Models for Subject-Driven Generation},
    author={Ruiz, Nataniel and Li, Yuanzhen and Jampani, Varun and Pritch, Yael and Rubinstein, Michael and Aberman, Kfir},
    booktitle={arXiv preprint arxiv:2208.12242},
    year={2022}
}

@article{ye2023ip-adapter,
    title={IP-Adapter: Text Compatible Image Prompt Adapter for Text-to-Image Diffusion Models},
    author={Ye, Hu and Zhang, Jun and Liu, Sibo and Han, Xiao and Yang, Wei},
    booktitle={arXiv preprint arxiv:2308.06721},
    year={2023}
}

@article{li2023blip,
    title={BLIP-Diffusion: Pre-trained Subject Representation for Controllable Text-to-Image Generation and Editing},
    author={Li, Dongxu and Li, Junnan and Hoi, Steven CH},
    journal={arXiv preprint arXiv:2305.14720},
    year={2023}
}

@article{Emu2,
    title={Generative Multimodal Models are In-Context Learners}, 
    author={Quan Sun and Yufeng Cui and Xiaosong Zhang and Fan Zhang and Qiying Yu and Zhengxiong Luo and Yueze Wang and Yongming Rao and Jingjing Liu and Tiejun Huang and Xinlong Wang},
    publisher={arXiv preprint arXiv:2312.13286},
    year={2023}
}

@misc{rombach2022highresolutionimagesynthesislatent,
      title={High-Resolution Image Synthesis with Latent Diffusion Models}, 
      author={Robin Rombach and Andreas Blattmann and Dominik Lorenz and Patrick Esser and Björn Ommer},
      year={2022},
      eprint={2112.10752},
      archivePrefix={arXiv},
      primaryClass={cs.CV},
      url={https://arxiv.org/abs/2112.10752}, 
}

@misc{li2022blipbootstrappinglanguageimagepretraining,
      title={BLIP: Bootstrapping Language-Image Pre-training for Unified Vision-Language Understanding and Generation}, 
      author={Junnan Li and Dongxu Li and Caiming Xiong and Steven Hoi},
      year={2022},
      eprint={2201.12086},
      archivePrefix={arXiv},
      primaryClass={cs.CV},
      url={https://arxiv.org/abs/2201.12086}, 
}

@misc{wang2024qwen2vlenhancingvisionlanguagemodels,
      title={Qwen2-VL: Enhancing Vision-Language Model's Perception of the World at Any Resolution}, 
      author={Peng Wang and Shuai Bai and Sinan Tan and Shijie Wang and Zhihao Fan and Jinze Bai and Keqin Chen and Xuejing Liu and Jialin Wang and Wenbin Ge and Yang Fan and Kai Dang and Mengfei Du and Xuancheng Ren and Rui Men and Dayiheng Liu and Chang Zhou and Jingren Zhou and Junyang Lin},
      year={2024},
      eprint={2409.12191},
      archivePrefix={arXiv},
      primaryClass={cs.CV},
      url={https://arxiv.org/abs/2409.12191}, 
}

@article{sun2021loftr,
  title={{LoFTR}: Detector-Free Local Feature Matching with Transformers},
  author={Sun, Jiaming and Shen, Zehong and Wang, Yuang and Bao, Hujun and Zhou, Xiaowei},
  journal={CVPR},
  year={2021}
}

@inproceedings{Edstedt_2024roma,
   title={RoMa: Robust Dense Feature Matching},
   url={http://dx.doi.org/10.1109/CVPR52733.2024.01871},
   DOI={10.1109/cvpr52733.2024.01871},
   booktitle={2024 IEEE/CVF Conference on Computer Vision and Pattern Recognition (CVPR)},
   publisher={IEEE},
   author={Edstedt, Johan and Sun, Qiyu and Bökman, Georg and Wadenbäck, Mårten and Felsberg, Michael},
   year={2024},
   month=June, pages={19790–19800}
}

@misc{leroy2024groundingimagematching3dmast3r,
      title={Grounding Image Matching in 3D with MASt3R}, 
      author={Vincent Leroy and Yohann Cabon and Jérôme Revaud},
      year={2024},
      eprint={2406.09756},
      archivePrefix={arXiv},
      primaryClass={cs.CV},
      url={https://arxiv.org/abs/2406.09756}, 
}

@misc{lindenberger2023lightgluelocalfeaturematching,
      title={LightGlue: Local Feature Matching at Light Speed}, 
      author={Philipp Lindenberger and Paul-Edouard Sarlin and Marc Pollefeys},
      year={2023},
      eprint={2306.13643},
      archivePrefix={arXiv},
      primaryClass={cs.CV},
      url={https://arxiv.org/abs/2306.13643}, 
}

@article{ravi2024sam2,
  title={SAM 2: Segment Anything in Images and Videos},
  author={Ravi, Nikhila and Gabeur, Valentin and Hu, Yuan-Ting and Hu, Ronghang and Ryali, Chaitanya and Ma, Tengyu and Khedr, Haitham and R{\"a}dle, Roman and Rolland, Chloe and Gustafson, Laura and Mintun, Eric and Pan, Junting and Alwala, Kalyan Vasudev and Carion, Nicolas and Wu, Chao-Yuan and Girshick, Ross and Doll{\'a}r, Piotr and Feichtenhofer, Christoph},
  journal={arXiv preprint arXiv:2408.00714},
  url={https://arxiv.org/abs/2408.00714},
  year={2024}
}

@misc{ren2024grounded,
      title={Grounded SAM: Assembling Open-World Models for Diverse Visual Tasks}, 
      author={Tianhe Ren and Shilong Liu and Ailing Zeng and Jing Lin and Kunchang Li and He Cao and Jiayu Chen and Xinyu Huang and Yukang Chen and Feng Yan and Zhaoyang Zeng and Hao Zhang and Feng Li and Jie Yang and Hongyang Li and Qing Jiang and Lei Zhang},
      year={2024},
      eprint={2401.14159},
      archivePrefix={arXiv},
      primaryClass={cs.CV}
}

@InProceedings{lueddecke22_cvpr_clipseg,
    author    = {L\"uddecke, Timo and Ecker, Alexander},
    title     = {Image Segmentation Using Text and Image Prompts},
    booktitle = {Proceedings of the IEEE/CVF Conference on Computer Vision and Pattern Recognition (CVPR)},
    month     = {June},
    year      = {2022},
    pages     = {7086-7096}
}

@inproceedings{minderer2023scaling_owl,
  title={Scaling Open-Vocabulary Object Detection},
  author={Minderer, Matthias and Gritsenko, Alexey and Stone, Austin and Neumann, Maxim and Weissenborn, Dirk and Dosovitskiy, Alexey and Mahendran, Aravindh and Arnab, Anurag and Dehghani, Mostafa and Shen, Zirui and others},
  booktitle={Advances in Neural Information Processing Systems},
  volume={36},
  year={2023}
}

@misc{detone2018superpointselfsupervisedpointdetection,
      title={SuperPoint: Self-Supervised Interest Point Detection and Description}, 
      author={Daniel DeTone and Tomasz Malisiewicz and Andrew Rabinovich},
      year={2018},
      eprint={1712.07629},
      archivePrefix={arXiv},
      primaryClass={cs.CV},
      url={https://arxiv.org/abs/1712.07629}, 
}
